\documentclass[conference,final,]{IEEEtran}
\usepackage{cite}

\ifCLASSINFOpdf
\else
\fi
\usepackage{graphicx}

\usepackage{amsmath}
\usepackage[unicode=true]{hyperref}

\hypersetup{
            pdftitle={Endogenous Macrodynamics in Algorithmic Recourse},
            pdfkeywords={Algorithmic Recourse, Counterfactual Explanations, Explainable AI, Dynamic Systems},
            pdfborder={0 0 0},
            breaklinks=true}
\providecommand{\tightlist}{%
  \setlength{\itemsep}{0pt}\setlength{\parskip}{0pt}}

\usepackage{longtable,booktabs,array}
\usepackage{calc} 
\usepackage{etoolbox}
\makeatletter
\patchcmd\longtable{\par}{\if@noskipsec\mbox{}\fi\par}{}{}
\makeatother
\IfFileExists{footnotehyper.sty}{\usepackage{footnotehyper}}{\usepackage{footnote}}
\makesavenoteenv{longtable}

\newlength{\cslhangindent}
\newlength{\csllabelwidth}
\newlength{\cslentryspacingunit} 
  {}%
  {\par}
\newenvironment{CSLReferences}[2] 
 {
  \setlength{\parindent}{0pt}
  \ifodd #1
  \let\oldpar\par
  \def\par{\hangindent=\cslhangindent\oldpar}
  \fi
  \setlength{\parskip}{#2\cslentryspacingunit}
 }%
 {}
\usepackage{calc}

\newcommand{\CSLLeftMargin}[1]{\parbox[t]{\csllabelwidth}{#1}}
\newcommand{\CSLRightInline}[1]{\parbox[t]{\linewidth - \csllabelwidth}{#1}\break}

\IEEEoverridecommandlockouts
\usepackage{cite}
\usepackage{amsmath,amssymb,amsfonts}
\usepackage{algorithm}
\usepackage{algpseudocode}
\usepackage{graphicx}
\usepackage{textcomp}
\usepackage{xcolor}
\usepackage{placeins}
\def\BibTeX{{\rm B\kern-.05em{\sc i\kern-.025em b}\kern-.08em
    T\kern-.1667em\lower.7ex\hbox{E}\kern-.125emX}}
\usepackage{booktabs}
\usepackage{longtable}
\usepackage{array}
\usepackage{multirow}
\usepackage{wrapfig}
\usepackage{float}
\usepackage{colortbl}
\usepackage{pdflscape}
\usepackage{tabu}
\usepackage{threeparttable}
\usepackage{threeparttablex}
\usepackage[normalem]{ulem}
\usepackage{makecell}
\usepackage{xcolor}

\usepackage{amsthm}

\newtheorem{proposition}{Research Question}[section]

\theoremstyle{definition}

\theoremstyle{definition}
\newtheorem{example}{Example}[section]
\theoremstyle{definition}

\theoremstyle{definition}

\theoremstyle{remark}

\begin{document}
%
\title{Endogenous Macrodynamics in Algorithmic Recourse}


\author{



\IEEEauthorblockN{
Patrick Altmeyer
}
\IEEEauthorblockA{\emph{Delft University of Technology}\\
\emph{Faculty of Electrical Engineering}\\
\emph{Mathematics and Computer Science}\\
2628 XE Delft, The Netherlands
\\P.Altmeyer@tudelft.nl
}
\and
\IEEEauthorblockN{
Giovan Angela
}
\IEEEauthorblockA{\emph{Delft University of Technology}\\
\emph{Faculty of Electrical Engineering}\\
\emph{Mathematics and Computer Science}\\
2628 XE Delft, The Netherlands
\\G.J.A.Angela@student.tudelft.nl
}
\and
\IEEEauthorblockN{
Aleksander Buszydlik
}
\IEEEauthorblockA{\emph{Delft University of Technology}\\
\emph{Faculty of Electrical Engineering}\\
\emph{Mathematics and Computer Science}\\
2628 XE Delft, The Netherlands
\\A.J.Buszydlik@student.tudelft.nl
}
\and
\IEEEauthorblockN{
Karol Dobiczek
}
\IEEEauthorblockA{\emph{Delft University of Technology}\\
\emph{Faculty of Electrical Engineering}\\
\emph{Mathematics and Computer Science}\\
2628 XE Delft, The Netherlands
\\K.T.Dobiczek@student.tudelft.nl
}
\and
\IEEEauthorblockN{
Arie van Deursen
}
\IEEEauthorblockA{\emph{Delft University of Technology}\\
\emph{Faculty of Electrical Engineering}\\
\emph{Mathematics and Computer Science}\\
2628 XE Delft, The Netherlands
\\Arie.vanDeursen@tudelft.nl
}
\and
\IEEEauthorblockN{
Cynthia C. S. Liem
}
\IEEEauthorblockA{\emph{Delft University of Technology}\\
\emph{Faculty of Electrical Engineering}\\
\emph{Mathematics and Computer Science}\\
2628 XE Delft, The Netherlands
\\C.C.S.Liem@tudelft.nl
}

}


%


\maketitle

\begin{abstract}
Existing work on Counterfactual Explanations (CE) and Algorithmic Recourse (AR) has largely focused on single individuals in a static environment: given some estimated model, the goal is to find valid counterfactuals for an individual instance that fulfill various desiderata. The ability of such counterfactuals to handle dynamics like data and model drift remains a largely unexplored research challenge. There has also been surprisingly little work on the related question of how the actual implementation of recourse by one individual may affect other individuals. Through this work, we aim to close that gap. We first show that many of the existing methodologies can be collectively described by a generalized framework. We then argue that the existing framework does not account for a hidden external cost of recourse, that only reveals itself when studying the endogenous dynamics of recourse at the group level. Through simulation experiments involving various state-of-the-art counterfactual generators and several benchmark datasets, we generate large numbers of counterfactuals and study the resulting domain and model shifts. We find that the induced shifts are substantial enough to likely impede the applicability of Algorithmic Recourse in some situations. Fortunately, we find various strategies to mitigate these concerns. Our simulation framework for studying recourse dynamics is fast and open-sourced.
\end{abstract}



\maketitle


%
\IEEEpeerreviewmaketitle

\author{\IEEEauthorblockN{1\textsuperscript{st} Given Name Surname}
\IEEEauthorblockA{\textit{dept. name of organization (of Aff.)} \\
\textit{name of organization (of Aff.)}\\
City, Country \\
email address or ORCID}
\and
\IEEEauthorblockN{2\textsuperscript{nd} Given Name Surname}
\IEEEauthorblockA{\textit{dept. name of organization (of Aff.)} \\
\textit{name of organization (of Aff.)}\\
City, Country \\
email address or ORCID}
\and
\IEEEauthorblockN{3\textsuperscript{rd} Given Name Surname}
\IEEEauthorblockA{\textit{dept. name of organization (of Aff.)} \\
\textit{name of organization (of Aff.)}\\
City, Country \\
email address or ORCID}
\and
\IEEEauthorblockN{4\textsuperscript{th} Given Name Surname}
\IEEEauthorblockA{\textit{dept. name of organization (of Aff.)} \\
\textit{name of organization (of Aff.)}\\
City, Country \\
email address or ORCID}
\and
\IEEEauthorblockN{5\textsuperscript{th} Given Name Surname}
\IEEEauthorblockA{\textit{dept. name of organization (of Aff.)} \\
\textit{name of organization (of Aff.)}\\
City, Country \\
email address or ORCID}
\and
\IEEEauthorblockN{6\textsuperscript{th} Given Name Surname}
\IEEEauthorblockA{\textit{dept. name of organization (of Aff.)} \\
\textit{name of organization (of Aff.)}\\
City, Country \\
email address or ORCID}
}

\hypertarget{intro}{%
\section{Introduction}\label{intro}}

Recent advances in Artificial Intelligence (AI) have propelled its adoption in scientific domains outside of Computer Science including Healthcare, Bioinformatics, Genetics and the Social Sciences. While this has in many cases brought benefits in terms of efficiency, state-of-the-art models like Deep Neural Networks (DNN) have also given rise to a new type of problem in the context of data-driven decision-making. They are essentially \textbf{black boxes}: so complex, opaque and underspecified in the data that it is often impossible to understand how they actually arrive at their decisions without auxiliary tools. Despite this shortcoming, black-box models have grown in popularity in recent years and have at times created undesirable societal outcomes \protect\hyperlink{ref-oneil2016weapons}{{[}1{]}}. The scientific community has tackled this issue from two different angles: while some have appealed for a strict focus on inherently interpretable models \protect\hyperlink{ref-rudin2019stop}{{[}2{]}}, others have investigated different ways to explain the behaviour of black-box models. These two sub-domains can be broadly referred to as \textbf{interpretable AI} and \textbf{explainable AI} (XAI), respectively.

Among the approaches to XAI that have recently grown in popularity are \textbf{Counterfactual Explanations} (CE). They explain how inputs into a model need to change for it to produce different outputs. Counterfactual Explanations that involve realistic and actionable changes can be used for the purpose of \textbf{Algorithmic Recourse} (AR) to help individuals who face adverse outcomes. An example relevant to the Social Sciences is consumer credit: in this context, AR can be used to guide individuals in improving their creditworthiness, should they have previously been denied access to credit based on an automated decision-making system. A meaningful recourse recommendation for a denied applicant could be: \emph{``If your net savings rate had been 10\% of your monthly income instead of the actual 8\%, your application would have been successful. See if you can temporarily cut down on consumption.''} In the remainder of this paper, we will use both terminologies---recourse and counterfactual---interchangeably to refer to situations where counterfactuals are generated with the intent to provide individual recourse.

Existing work in this field has largely worked in a static setting: various approaches have been proposed to generate counterfactuals for a given individual that is subject to some pre-trained model. More recent work has compared different approaches within this static setting \protect\hyperlink{ref-pawelczyk2021carla}{{[}3{]}}. In this work, we go one step further and ask ourselves: what happens if recourse is provided and implemented repeatedly? What types of dynamics are introduced and how do different counterfactual generators compare in this context?

Research on Algorithmic Recourse has also so far typically addressed the issue from the perspective of a single individual. Arguably though, most real-world applications that warrant AR involve potentially large groups of individuals typically competing for scarce resources. Our work demonstrates that in such scenarios, choices made by or for a single individual are likely to affect the broader collective of individuals in ways that many current approaches to AR fail to account for. More specifically, we argue that a strict focus on minimizing the private costs to individuals may be too narrow an objective.

\begin{figure}

{\centering \includegraphics[width=0.9\linewidth]{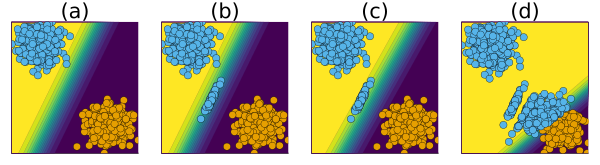} 

}

\caption{Dynamics in Algorithmic Recourse: (a) we have a simple linear classifier trained for binary classification where samples from the negative class ($y=0$) are marked in orange and samples of the positive class ($y=1$) are marked in blue; (b) the implementation of AR for a random subset of individuals leads to a noticeable domain shift; (c) as the classifier is retrained we observe a corresponding model shift; (d) as this process is repeated, the decision boundary moves away from the target class.}\label{fig:poc}
\end{figure}

Figure \ref{fig:poc} illustrates this idea for a binary problem involving a linear classifier and the counterfactual generator proposed by Wachter et al. \protect\hyperlink{ref-wachter2017counterfactual}{{[}4{]}}: the implementation of AR for a subset of individuals immediately leads to a visible domain shift in the (blue) target class (b), which in turn triggers a model shift (c). As this game of implementing AR and updating the classifier is repeated, the decision boundary moves away from training samples that were originally in the target class (d). We refer to these types of dynamics as \textbf{endogenous} because they are induced by the implementation of recourse itself. The term \textbf{macrodynamics} is borrowed from the economics literature and used to describe processes involving whole groups or societies.

We think that these types of endogenous dynamics may be problematic and deserve our attention. From a purely technical perspective, we note the following: firstly, model shifts may inadvertently change classification outcomes for individuals who never received recourse. Secondly, we observe in Figure \ref{fig:poc} that as the decision boundary moves in the direction of the non-target class, counterfactual paths become shorter. We think that in some practical applications, this can be expected to generate costs for involved stakeholders. To follow our argument, consider the following two examples:

\begin{example}[Consumer Credit]
\protect\hypertarget{exm:consumer}{}\label{exm:consumer}Suppose Figure \ref{fig:poc} relates to an automated decision-making system used by a retail bank to evaluate credit applicants with respect to their creditworthiness. Assume that the two features are meaningful in the sense that creditworthiness decreases in the bottom-right direction. Then we can think of the outcome in panel (d) as representing a situation where the bank supplies credit to more borrowers (blue), but these borrowers are on average less creditworthy and more of them can be expected to default on their loan. This represents a cost to the retail bank.
\end{example}

\begin{example}[Student Admission]
\protect\hypertarget{exm:student}{}\label{exm:student}Suppose Figure \ref{fig:poc} relates to an automated decision-making system used by a university in its student admission process. Assume that the two features are meaningful in the sense that the likelihood of students completing their degree decreases in the bottom-right direction. Then we can think of the outcome in panel (b) as representing a situation where more students are admitted to university (blue), but they are more likely to fail their degree than students that were admitted in previous years. The university admission committee catches on to this and suspends its efforts to offer Algorithmic Recourse. This represents an opportunity cost to future student applicants, that may have derived utility from being offered recourse.
\end{example}

Both examples are exaggerated simplifications of potential real-world scenarios, but they serve to illustrate the point that recourse for one single individual may exert negative externalities on other individuals.

To the best of our knowledge, this is the first work investigating endogenous macrodynamics in AR. Our contributions to the state of knowledge are as follows: firstly, we posit a compelling argument that calls for a novel perspective on Algorithmic Recourse extending our focus from single individuals to groups (Sections \ref{related} and \ref{method}). Secondly, we introduce an experimental framework extending previous work by Altmeyer \protect\hyperlink{ref-altmeyer2022counterfactualexplanations}{{[}5{]}}, which enables us to study macrodynamics of Algorithmic Recourse through simulations that can be fully parallelized (Section \ref{method-2}). Thirdly, we use this framework to provide a first in-depth analysis of endogenous recourse dynamics induced by various popular counterfactual generators proposed in \protect\hyperlink{ref-wachter2017counterfactual}{{[}4{]}}, \protect\hyperlink{ref-schut2021generating}{{[}6{]}}, \protect\hyperlink{ref-joshi2019realistic}{{[}7{]}}, \protect\hyperlink{ref-mothilal2020explaining}{{[}8{]}} and \protect\hyperlink{ref-antoran2020getting}{{[}9{]}} (Sections \ref{empirical} and \ref{empirical-2}). Fourthly, given that we find a substantial impact of recourse, we propose and assess various mitigation strategies (Section \ref{mitigate}). Finally, we discuss our findings in the broader context of the literature in Section \ref{discussion}, before pointing to some of the limitations of our work as well as avenues for future research in Section \ref{limit}. Section \ref{conclusion} concludes.

\hypertarget{related}{%
\section{Background}\label{related}}

In this section, we provide a review of the relevant literature. First, Subsection \ref{related-recourse} discusses the existing research within the domain of Counterfactual Explanations and Algorithmic Recourse. Then, Subsection \ref{related-shifts} presents some of the previous work on the measurement of data and model shifts.

\hypertarget{related-recourse}{%
\subsection{Algorithmic Recourse}\label{related-recourse}}

A framework for Counterfactual Explanations was first proposed in 2017 by Wachter et al. \protect\hyperlink{ref-wachter2017counterfactual}{{[}4{]}} and has served as the baseline for many methodologies that have been proposed since then. Let \(M: \mathcal{X} \mapsto \mathcal{Y}\) denote some pre-trained model that maps from inputs \(X \in \mathcal{X}\) to outputs \(Y \in \mathcal{Y}\). Then we are interested in minimizing the cost\footnote{Equivalently, others have referred to this quantity as \emph{complexity} or simply \emph{distance}.} \(C=\text{cost}(x^\prime)\) incurred by individual \(x\) when moving to a counterfactual state \(x^\prime\) such that the predicted outcome \(M(x^\prime)\) corresponds to some target outcome \(y^*\):

\begin{equation}
\min_{x^\prime \in \mathcal{X}} \text{cost}(x^\prime) \ \ \ \mbox{s. t.} \ \ \ M(x^\prime) = y^* \label{eq:obj}
\end{equation}

For implementation purposes, \eqref{eq:obj} is typically approximated through regularization:

\begin{equation}
x^\prime = \arg \min_{x^\prime}  \text{yloss}(M(x^\prime),y^*) + \lambda \text{cost}(x^\prime) \label{eq:solution}
\end{equation}

In the baseline work \protect\hyperlink{ref-wachter2017counterfactual}{{[}4{]}}, the cost function is proxied by some distance metric based on the simple intuition that perturbations of \(x\) are costly to the individual. For models that are differentiable and produce smooth predictions, \eqref{eq:solution} can be solved through gradient descent. This summarizes the approach followed in \protect\hyperlink{ref-wachter2017counterfactual}{{[}4{]}} which we refer to simply as \textbf{Wachter}, the name of the first author, in the remainder of this paper.

Many approaches for the generation of Algorithmic Recourse have been described in the literature since 2017. An October 2020 survey by Karimi et al.~laid out 60 algorithms that have been proposed since 2014 \protect\hyperlink{ref-karimi2020survey}{{[}10{]}}. Another survey published around the same time by Verma et al.~described 29 algorithms \protect\hyperlink{ref-verma2020counterfactual}{{[}11{]}}. Different approaches vary primarily in terms of the objective functions they impose, how they optimize said objective (from brute force through gradient-based approaches to graph traversal algorithms), and how they ensure that certain requirements for CE are met. Regarding the latter, the literature has produced an extensive list of desiderata each addressing different needs. To name but a few, we are interested in generating counterfactuals that are close \protect\hyperlink{ref-wachter2017counterfactual}{{[}4{]}}, actionable \protect\hyperlink{ref-ustun2019actionable}{{[}12{]}}, realistic \protect\hyperlink{ref-schut2021generating}{{[}6{]}}, sparse, diverse \protect\hyperlink{ref-mothilal2020explaining}{{[}8{]}} and if possible causally founded \protect\hyperlink{ref-karimi2021algorithmic}{{[}13{]}}.

Efforts so far have largely been directed at improving the quality of Counterfactual Explanations within a static context: given some pre-trained classifier \(M: \mathcal{X} \mapsto \mathcal{Y}\), we are interested in generating one or multiple meaningful Counterfactual Explanations for some individual characterized by \(x\). The ability of Counterfactual Explanations to handle dynamics like data and model shifts remains a largely unexplored research challenge at this point \protect\hyperlink{ref-verma2020counterfactual}{{[}11{]}}. We have been able to identify only one recent work by Upadhyay et al.~that considers the implications of \textbf{exogenous} domain and model shifts in the context of AR \protect\hyperlink{ref-upadhyay2021robust}{{[}14{]}}. Exogenous shifts are strictly of external origin. For example, they might stem from data correction, temporal shifts or geospatial changes \protect\hyperlink{ref-upadhyay2021robust}{{[}14{]}}. Upadhyay et al. \protect\hyperlink{ref-upadhyay2021robust}{{[}14{]}} propose ROAR: a framework for Algorithmic Recourse that evidently improves robustness to such exogenous shifts.

As mentioned earlier, research has so far also generally focused on generating counterfactuals for single individuals or instances. We have been able to identify only one existing work that investigates black-box model behaviour towards a group of individuals \protect\hyperlink{ref-carrizosa2021generating}{{[}15{]}}. The authors propose an optimization framework that generates collective counterfactuals. We provide a motivation for doing so from the perspective of endogenous macrodynamics of Algorithmic Recourse.

\hypertarget{related-shifts}{%
\subsection{Domain and Model Shifts}\label{related-shifts}}

Much attention has been paid to the detection of dataset shifts -- situations where the distribution of data changes over time. Rabanser et al.~suggest a framework to detect data drift from a minimal number of samples through the application of two-sample tests \protect\hyperlink{ref-rabanser2019failing}{{[}16{]}}. This task is a generalization of the anomaly detection problem for large datasets, which aims to answer the question if two sets of samples could have been generated from the same probability distribution. Numerous approaches to anomaly detection have been summarized \protect\hyperlink{ref-chandola2009anomaly}{{[}17{]}}. Another well-established research topic is concept drift: situations where external variables influence the patterns between the input and the output of a model \protect\hyperlink{ref-widmer1996learning}{{[}18{]}}. For instance, Gama et al.~offer a review of the adaptive learning techniques which can handle concept drift \protect\hyperlink{ref-gama2014survey}{{[}19{]}}. Less previous work is available on the related topic of model drift: changes in model performance over time. Nelson et al.~review how resistant different machine learning models are to model drift \protect\hyperlink{ref-nelson2015evaluating}{{[}20{]}}. Ackerman et al.~offer a method to detect changes in model performance when ground truth is not available \protect\hyperlink{ref-ackerman2021machine}{{[}21{]}}.

In the context of Algorithmic Recourse, domain and model shifts were first brought up by the authors behind ROAR \protect\hyperlink{ref-upadhyay2021robust}{{[}14{]}}. In their work, they refer to model shifts as simply any perturbation \(\Delta\) to the parameters of the model in question: \(M\). While this also sets the baseline for our analysis here, it is worth noting that in \protect\hyperlink{ref-upadhyay2021robust}{{[}14{]}} these perturbations are mechanically introduced. In contrast, we are interested in quantifying model shifts that arise endogenously as part of a dynamic recourse process. In addition to quantifying the magnitude of shifts \(\Delta\), we aim to also analyse the characteristics of changes to the model, such as the position of the decision boundary and the overall decisiveness of the model. We have not been able to identify previous work on this topic.

\hypertarget{related-benchmark}{%
\subsection{Benchmarking Counterfactual Generators}\label{related-benchmark}}

Despite the large and growing number of approaches to counterfactual search, there have been surprisingly few benchmark studies that compare different methodologies. This may be partially due to limited software availability in this space. Recent work has started to address this gap: firstly, \protect\hyperlink{ref-deoliveira2021framework}{{[}22{]}} run a large benchmarking study using different algorithmic approaches and numerous tabular datasets; secondly, \protect\hyperlink{ref-pawelczyk2021carla}{{[}3{]}} introduce a Python framework---CARLA---that can be used to apply and benchmark different methodologies; finally, \href{https://github.com/pat-alt/CounterfactualExplanations.jl}{\texttt{CounterfactualExplanations.jl}} \protect\hyperlink{ref-altmeyer2022counterfactualexplanations}{{[}5{]}} provides an extensible and fast implementation in Julia. Since the experiments presented here involve extensive simulations, we have relied on and extended the Julia implementation due to the associated performance benefits. In particular, we have built a framework on top of \href{https://github.com/pat-alt/CounterfactualExplanations.jl}{\texttt{CounterfactualExplanations.jl}} that extends the functionality from static benchmarks to simulation experiments: \href{(https://github.com/pat-alt/AlgorithmicRecourseDynamics.jl)}{\texttt{AlgorithmicRecourseDynamics.jl}}\footnote{The code has been released as a package: \url{https://github.com/pat-alt/AlgorithmicRecourseDynamics.jl}.}. The core concepts implemented in that package reflect what is presented in Section \ref{method-2} of this paper.

\hypertarget{method}{%
\section{Gradient-Based Recourse Revisited}\label{method}}

In this section, we first set out a generalized framework for gradient-based counterfactual search that encapsulates the various Individual Recourse methods we have chosen to use in our experiments (Section \ref{method-general}). We then introduce the notion of a hidden external cost in Algorithmic Recourse and extend the existing framework to explicitly address this cost in the counterfactual search objective (Section \ref{method-collective}).

\hypertarget{method-general}{%
\subsection{From Individual Recourse \ldots{}}\label{method-general}}

We have chosen to focus on gradient-based counterfactual search for two reasons: firstly, they can be seen as direct descendants of our baseline method (Wachter); secondly, gradient-based search is particularly well-suited for differentiable black-box models like deep neural networks, which we focus on in this work. In particular, we include the following generators in our simulation experiments below: \textbf{REVISE} \protect\hyperlink{ref-joshi2019realistic}{{[}7{]}}, \textbf{CLUE} \protect\hyperlink{ref-antoran2020getting}{{[}9{]}}, \textbf{DiCE} \protect\hyperlink{ref-mothilal2020explaining}{{[}8{]}} and a greedy approach that relies on probabilistic models \protect\hyperlink{ref-schut2021generating}{{[}6{]}}. Our motivation for including these different generators in our analysis is that they all offer slightly different approaches to generating meaningful counterfactuals for differentiable black-box models. We hypothesize that generating more \textbf{meaningful} counterfactuals should mitigate the endogenous dynamics illustrated in Figure \ref{fig:poc} in Section \ref{intro}. This intuition stems from the underlying idea that more meaningful counterfactuals are generated by the same or at least a very similar data-generating process as the observed data. All else equal, counterfactuals that fulfil this basic requirement should be less prone to trigger shifts.

As we will see next, all of them can be described by the following generalized form of Equation \eqref{eq:general}:

\begin{equation}
\begin{aligned}
\mathbf{s}^\prime &= \arg \min_{\mathbf{s}^\prime \in \mathcal{S}} \left\{  {\text{yloss}(M(f(\mathbf{s}^\prime)),y^*)}+ \lambda {\text{cost}(f(\mathbf{s}^\prime)) }  \right\} \label{eq:general}
\end{aligned} 
\end{equation}

Here \(\mathbf{s}^\prime=\left\{s_k^\prime\right\}_K\) is a \(K\)-dimensional array of counterfactual states and \(f: \mathcal{S} \mapsto \mathcal{X}\) maps from the counterfactual state space to the feature space. In Wachter, the state space is the feature space: \(f\) is the identity function and the number of counterfactuals \(K\) is one. Both REVISE and CLUE search counterfactuals in some latent space \(\mathcal{S}\) instead of the feature space directly. The latent embedding is learned by a separate generative model that is tasked with learning the data-generating process (DGP) of \(X\). In this case, \(f\) in Equation \eqref{eq:general} corresponds to the decoder part of the generative model, that is the function that maps back from the latent space to inputs. Provided the generative model is well-specified, traversing the latent embedding typically yields meaningful counterfactuals since they are implicitly generated by the (learned) DGP \protect\hyperlink{ref-joshi2019realistic}{{[}7{]}}.

CLUE distinguishes itself from REVISE and other counterfactual generators in that it aims to minimize the predictive uncertainty of the model in question, \(M\). To quantify predictive uncertainty, Antoran et al. \protect\hyperlink{ref-antoran2020getting}{{[}9{]}} rely on entropy estimates for probabilistic models. The greedy approach proposed by Schut et al. \protect\hyperlink{ref-schut2021generating}{{[}6{]}}, which we refer to as \textbf{Greedy}, also works with the subclass of models \(\tilde{\mathcal{M}}\subset\mathcal{M}\) that can produce predictive uncertainty estimates. The authors show that in this setting the cost function \(\text{cost}(\cdot)\) in Equation \eqref{eq:general} is redundant and meaningful counterfactuals can be generated in a fast and efficient manner through a modified Jacobian-based Saliency Map Attack (JSMA). Schut et al. \protect\hyperlink{ref-schut2021generating}{{[}6{]}} also show that by maximizing the predicted probability of \(x^\prime\) being assigned to target class \(y^*\), we also implicitly minimize predictive entropy (as in CLUE). In that sense, CLUE can be seen as equivalent to REVISE in the Bayesian context and we shall therefore refer to both approaches collectively as \textbf{Latent Space} generators\footnote{In fact, there are several other recently proposed approaches to counterfactual search that also broadly fall in this same category. They largely differ with respect to the chosen generative model: for example, the generator proposed by Dombrowski et al. \protect\hyperlink{ref-dombrowski2021diffeomorphic}{{[}23{]}} relies on normalizing flows.}.

Finally, DiCE \protect\hyperlink{ref-mothilal2020explaining}{{[}8{]}} distinguishes itself from all other generators considered here in that it aims to generate a diverse set of \(K>1\) counterfactuals. Wachter et al. \protect\hyperlink{ref-wachter2017counterfactual}{{[}4{]}} show that diverse outcomes can in principle be achieved simply by rerunning counterfactual search multiple times using stochastic gradient descent (or by randomly initializing the counterfactual)\footnote{Note that \eqref{eq:general} naturally lends itself to that idea: setting \(K\) to some value greater than one and using the Wachter objective essentially boils down to computing multiple counterfactuals in parallel. Here, \(yloss(\cdot)\) is first broadcasted over elements of \(\mathbf{s}^\prime\) and then aggregated. This is exactly how counterfactual search is implemented in \href{https://github.com/pat-alt/CounterfactualExplanations.jl}{\texttt{CounterfactualExplanations.jl}}.}. In \protect\hyperlink{ref-mothilal2020explaining}{{[}8{]}} diversity is explicitly proxied via Determinantal Point Processes (DDP): the authors introduce DDP as a component of the cost function \(\text{cost}(\mathbf{s}^\prime)\) and thereby produce counterfactuals \(s_1, ..., s_K\) that look as different from each other as possible. The implementation of DiCE in our library of choice---\href{https://github.com/pat-alt/CounterfactualExplanations.jl}{\texttt{CounterfactualExplanations.jl}}---uses that exact approach. It is worth noting that for \(k=1\), DiCE reduces to Wachter since the DDP is constant and therefore does not affect the objective function in Equation \eqref{eq:general}.

\hypertarget{method-collective}{%
\subsection{\ldots{} towards Collective Recourse}\label{method-collective}}

All of the different approaches introduced above tackle the problem of Algorithmic Recourse from the perspective of one single individual\footnote{DiCE recognizes that different individuals may have different objective functions, but it does not address the interdependencies between different individuals.}. To explicitly address the issue that Individual Recourse may affect the outcome and prospect of other individuals, we propose to extend Equation \eqref{eq:general} as follows:

\begin{equation}
\begin{aligned}
\mathbf{s}^\prime &= \arg \min_{\mathbf{s}^\prime \in \mathcal{S}} \{ {\text{yloss}(M(f(\mathbf{s}^\prime)),y^*)} \\ &+ \lambda_1 {\text{cost}(f(\mathbf{s}^\prime))} + \lambda_2 {\text{extcost}(f(\mathbf{s}^\prime))} \}  \label{eq:collective}
\end{aligned} 
\end{equation}

Here \(\text{cost}(f(\mathbf{s}^\prime))\) denotes the proxy for private costs faced by the individual as before and \(\lambda_1\) governs to what extent that private cost ought to be penalized. The newly introduced term \(\text{extcost}(f(\mathbf{s}^\prime))\) is meant to capture and address external costs incurred by the collective of individuals in response to changes in \(\mathbf{s}^\prime\). The underlying concept of private and external costs is borrowed from Economics and well-established in that field: when the decisions or actions by some individual market participant generate external costs, then the market is said to suffer from negative externalities and is considered inefficient \protect\hyperlink{ref-pindyck2014microeconomics}{{[}24{]}}. We think that this concept describes the endogenous dynamics of algorithmic recourse observed here very well. As with Individual Recourse, the exact choice of \(\text{extcost}(\cdot)\) is not obvious, nor do we intend to provide a definitive answer in this work, if such even exists. That being said, we do propose a few potential mitigation strategies in Section \ref{mitigate}.

\hypertarget{method-2}{%
\section{Modelling Endogenous Macrodynamics in Algorithmic Recourse}\label{method-2}}

In the following, we describe the framework we propose for modelling and analyzing endogenous macrodynamics in Algorithmic Recourse. We introduce this framework with the ambition to shed light on the following research questions:

\begin{proposition}[Endogenous Shifts]
\protect\hypertarget{prp:shifts}{}\label{prp:shifts}Does the repeated implementation of recourse provided by state-of-the-art generators lead to shifts in the domain and model?
\end{proposition}

\begin{proposition}[Costs]
\protect\hypertarget{prp:costs}{}\label{prp:costs}If so, are these dynamics substantial enough to be considered costly to stakeholders involved in real-world automated decision-making processes?
\end{proposition}

\begin{proposition}[Heterogeneity]
\protect\hypertarget{prp:het}{}\label{prp:het}Do different counterfactual generators yield significantly different outcomes in this context? Furthermore, is there any heterogeneity concerning the chosen classifier and dataset?
\end{proposition}

\begin{proposition}[Drivers]
\protect\hypertarget{prp:drive}{}\label{prp:drive}What are the drivers of endogenous dynamics in Algorithmic Recourse?
\end{proposition}

Below we first describe the basic simulations that were generated to produce the findings in this work and also constitute the core of \href{https://anonymous.4open.science/r/AlgorithmicRecourseDynamics/README.md}{\texttt{AlgorithmicRecourseDynamics.jl}}---the Julia package we introduced earlier. The remainder of this section then introduces various evaluation metrics that can be used to benchmark different counterfactual generators with respect to how they perform in the dynamic setting.

\hypertarget{method-2-experiment}{%
\subsection{Simulations}\label{method-2-experiment}}

The dynamics illustrated in Figure \ref{fig:poc} were generated through a simple experiment that aims to simulate the process of Algorithmic Recourse in practice. We begin in the static setting at time \(t=0\): firstly, we have some binary classifier \(M\) that was pre-trained on data \(\mathcal{D}=\mathcal{D}_0 \cup \mathcal{D}_1\), where \(\mathcal{D}_0\) and \(\mathcal{D}_1\) denote samples in the non-target and target class, respectively; secondly, we generate recourse for a random batch of \(B\) individuals in the non-target class (\(\mathcal{D}_0\)). Note that we focus our attention on classification problems since classification poses the most common use-case for recourse\footnote{To keep notation simple, we have also restricted ourselves to binary classification here, but \texttt{AlgorithmicRecourseDynamics.jl} can also be used for multi-class problems.}.

In order to simulate the dynamic process, we suppose that the model \(M\) is retrained following the actual implementation of recourse in time \(t=0\). Following the update to the model, we assume that at time \(t=1\) recourse is generated for yet another random subset of individuals in the non-target class. This process is repeated for a number of time periods \(T\). To get a clean read on endogenous dynamics we keep the total population of samples closed: we allow existing samples to move from factual to counterfactual states but do not allow any entirely new samples to enter the population. The experimental setup is summarized in Algorithm \ref{algo-experiment}.

\begin{algorithm}
\caption{Simulation Experiment}\label{algo-experiment}
\begin{algorithmic}[1]
\Procedure{Experiment}{$M,\mathcal{D},G$}
\State $E\gets \emptyset$ \Comment{Initialize evaluation $E$.}
\State $t\gets 0$
\While{$t<T$}
\State $\text{batch} \subset \mathcal{D}_0$ \Comment{Sample from $\mathcal{D}_0$  (assignment).}
\State $\text{batch}\gets G(\text{batch})$ \Comment{Generate counterfactuals.}
\State $M\gets M(\mathcal{D})$ \Comment{Retrain model.}
\State $E\gets \text{eval}(M,\mathcal{D}) \cup E$ \Comment{Update evaluation.}
\State $t\gets t+1$ \Comment{Increment $t$.}
\EndWhile
\State \textbf{return} $E, M,\mathcal{D}$
\EndProcedure
\end{algorithmic}
\end{algorithm}

Note that the operation in line 4 is an assignment, rather than a copy operation, so any updates to `batch' will also affect \(\mathcal{D}\). The function \(\text{eval}(M,\mathcal{D})\) loosely denotes the computation of various evaluation metrics introduced below. In practice, these metrics can also be computed at regular intervals as opposed to every round.

Along with any other fixed parameters affecting the counterfactual search, the parameters \(T\) and \(B\) are assumed as given in Algorithm \ref{algo-experiment}. Still, it is worth noting that the higher these values, the more factual instances undergo recourse throughout the entire experiment. Of course, this is likely to lead to more pronounced domain and model shifts by time \(T\). In our experiments, we choose the values such that the majority of the negative instances from the initial dataset receive recourse. As we compute evaluation metrics at regular intervals throughout the procedure, we can also verify the impact of recourse when it is implemented for a smaller number of individuals.

Algorithm \ref{algo-experiment} summarizes the proposed simulation experiment for a given dataset \(\mathcal{D}\), model \(M\) and generator \(G\), but naturally, we are interested in comparing simulation outcomes for different sources of data, models and generators. The framework we have built facilitates this, making use of multi-threading in order to speed up computations. Holding the initial model and dataset constant, the experiments are run for all generators, since our primary concern is to benchmark different recourse methods. To ensure that each generator is faced with the same initial conditions in each round \(t\), the candidate batch of individuals from the non-target class is randomly drawn from the intersection of all non-target class individuals across all experiments \(\left\{\textsc{Experiment}(M,\mathcal{D},G)\right\}_{j=1}^J\) where \(J\) is the total number of generators.

\hypertarget{method-2-metrics}{%
\subsection{Evaluation Metrics}\label{method-2-metrics}}

We formulate two desiderata for the set of metrics used to measure domain and model shifts induced by recourse. First, the metrics should be applicable regardless of the dataset or classification technique so that they allow for the meaningful comparison of the generators in various scenarios. As knowledge of the underlying probability distribution is rarely available, the metrics should be empirical and non-parametric. This further ensures that we can also measure large datasets by sampling from the available data. Moreover, while our study was conducted in a two-class classification setting, our choice of metrics should remain applicable in future research on multi-class recourse problems. Second, the set of metrics should allow capturing various aspects of the previously mentioned magnitude, path, and pace of changes while remaining as small as possible.

\hypertarget{domain-shifts}{%
\subsubsection{Domain Shifts}\label{domain-shifts}}

To quantify the magnitude of domain shifts we rely on an unbiased estimate of the squared population \textbf{Maximum Mean Discrepancy (MMD)} given as:

\begin{equation}
\begin{aligned}
MMD({X}^\prime,\tilde{X}^\prime) &= \frac{1}{m(m-1)}\sum_{i=1}^m\sum_{j\neq i}^m k(x_i,x_j) \\ &+ \frac{1}{n(n-1)}\sum_{i=1}^n\sum_{j\neq i}^n k(\tilde{x}_i,\tilde{x}_j) \\ &- \frac{2}{mn}\sum_{i=1}^m\sum_{j=1}^n k(x_i,\tilde{x}_j) \label{eq:mmd}
\end{aligned}
\end{equation}

where \(X=\{x_1,...,x_m\}\), \(\tilde{X}=\{\tilde{x}_1,...,\tilde{x}_n\}\) represent independent and identically distributed samples drawn from probability distributions \(\mathcal{X}\) and \(\mathcal{\tilde{X}}\) respectively \protect\hyperlink{ref-gretton2012kernel}{{[}25{]}}. MMD is a measure of the distance between the kernel mean embeddings of \(\mathcal{X}\) and \(\mathcal{\tilde{X}}\) in a Reproducing Kernel Hilbert Space, \(\mathcal{H}\) \protect\hyperlink{ref-berlinet2011reproducing}{{[}26{]}}. An important consideration is the choice of the kernel function \(k(\cdot,\cdot)\). In our implementation, we make use of a Gaussian kernel with a constant length-scale parameter of \(0.5\). As the Gaussian kernel captures all moments of distributions \(\mathcal{X}\) and \(\mathcal{\tilde{X}}\), we have that \(MMD(X,\tilde{X})=0\) if and only if \(X=\tilde{X}\). Conversely, larger values \(MMD(X,\tilde{X})>0\) indicate that it is more likely that \(\mathcal{X}\) and \(\mathcal{\tilde{X}}\) are different distributions. In our context, large values, therefore, indicate that a domain shift indeed seems to have occurred.

To assess the statistical significance of the observed shifts under the null hypothesis that samples \(X\) and \(\tilde{X}\) were drawn from the same probability distribution, we follow \protect\hyperlink{ref-arcones1992bootstrap}{{[}27{]}}. To that end, we combine the two samples and generate a large number of permutations of \(X + \tilde{X}\). Then, we split the permuted data into two new samples \(X^\prime\) and \(\tilde{X}^\prime\) having the same size as the original samples. Under the null hypothesis, we should have that \(MMD(X^\prime,\tilde{X}^\prime)\) be approximately equal to \(MMD(X,\tilde{X})\). The corresponding \(p\)-value can then be calculated by counting how often these two quantities are not equal.

We calculate the MMD for both classes individually based on the ground truth labels, i.e.~the labels that samples were assigned in time \(t=0\). Throughout our experiments, we generally do not expect the distribution of the negative class to change over time -- application of recourse reduces the size of this class, but since individuals are sampled uniformly the distribution should remain unaffected. Conversely, unless a recourse generator can perfectly replicate the original probability distribution, we expect the MMD of the positive class to increase. Thus, when discussing MMD, we generally mean the shift in the distribution of the positive class.

\hypertarget{model-shifts}{%
\subsubsection{Model Shifts}\label{model-shifts}}

As our baseline for quantifying model shifts, we measure perturbations to the model parameters at each point in time \(t\) following \protect\hyperlink{ref-upadhyay2021robust}{{[}14{]}}. We define \(\Delta=||\theta_{t+1}-\theta_{t}||^2\), that is the euclidean distance between the vectors of parameters before and after retraining the model \(M\). We shall refer to this baseline metric simply as \textbf{Perturbations}.

We extend the metric in Equation \eqref{eq:mmd} to quantify model shifts. Specifically, we introduce \textbf{Predicted Probability MMD (PP MMD)}: instead of applying Equation \eqref{eq:mmd} to features directly, we apply it to the predicted probabilities assigned to a set of samples by the model \(M\). If the model shifts, the probabilities assigned to each sample will change; again, this metric will equal 0 only if the two classifiers are the same. We compute PP MMD in two ways: firstly, we compute it over samples drawn uniformly from the dataset, and, secondly, we compute it over points spanning a mesh grid over a subspace of the entire feature space. For the latter approach, we bound the subspace by the extrema of each feature. While this approach is theoretically more robust, unfortunately, it suffers from the curse of dimensionality, since it becomes increasingly difficult to select enough points to overcome noise as the dimension \(D\) grows.

As an alternative to PP MMD, we use a pseudo-distance for the \textbf{Disagreement Coefficient} (Disagreement). This metric was introduced in \protect\hyperlink{ref-hanneke2007bound}{{[}28{]}} and estimates \(p(M(x) \neq M^\prime(x))\), that is the probability that two classifiers disagree on the predicted outcome for a randomly chosen sample. Thus, it is not relevant whether the classification is correct according to the ground truth, but only whether the sample lies on the same side of the two respective decision boundaries. In our context, this metric quantifies the overlap between the initial model (trained before the application of AR) and the updated model. A Disagreement Coefficient unequal to zero is indicative of a model shift. The opposite is not true: even if the Disagreement Coefficient is equal to zero, a model shift may still have occurred. This is one reason why PP MMD is our preferred metric.

We further introduce \textbf{Decisiveness} as a metric that quantifies the likelihood that a model assigns a high probability to its classification of any given sample. We define the metric simply as \({\frac{1}{N}}\sum_{i=0}^N(\sigma(M(x)) - 0.5)^2\) where \(M(x)\) are predicted logits from a binary classifier and \(\sigma\) denotes the sigmoid function. This metric provides an unbiased estimate of the binary classifier's tendency to produce high-confidence predictions in either one of the two classes. Although the exact values for this metric are not important for our study, they can be used to detect model shifts. If decisiveness changes over time, then this is indicative of the decision boundary moving towards either one of the two classes. A potential caveat of this metric in the context of our experiments is that it will to some degree get inflated simply through retraining the model.

Finally, we also take a look at the out-of-sample \textbf{Performance} of our models. To this end, we compute their F-score on a test sample that we leave untouched throughout the experiment.

\hypertarget{empirical}{%
\section{Experiment Setup}\label{empirical}}

This section presents the exact ingredients and parameter choices describing the simulation experiments we ran to produce the findings presented in the next section (\ref{empirical-2}). For convenience, we use Algorithm \ref{algo-experiment} as a template to guide us through this section. A few high-level details upfront: each experiment is run for a total of \(T=50\) rounds, where in each round we provide recourse to five per cent of all individuals in the non-target class, so \(B_t=0.05 * N_t^{\mathcal{D}_0}\). All classifiers and generative models are retrained for 10 epochs in each round \(t\) of the experiment. Rather than retraining models from scratch, we initialize all parameters at their previous levels (\(t-1\)) and backpropagate for 10 epochs using the new training data as inputs into the existing model. Evaluation metrics are computed and stored every 10 rounds. To account for noise, each individual experiment is repeated five times.\footnote{In the current implementation, we use the same train-test split each time to only account for stochasticity associated with randomly selecting individuals for recourse. An interesting alternative may be to also perform data splitting each time, thereby adding an additional layer of randomness.}

\hypertarget{empirical-classifiers}{%
\subsection{\texorpdfstring{\(M\)---Classifiers and Generative Models}{M---Classifiers and Generative Models}}\label{empirical-classifiers}}

For each dataset and generator, we look at three different types of classifiers, all of them built and trained using \texttt{Flux.jl} \protect\hyperlink{ref-innes2018fashionable}{{[}29{]}}: firstly, a simple linear classifier---\textbf{Logistic Regression}---implemented as a single linear layer with sigmoid activation; secondly, a multilayer perceptron (\textbf{MLP}); and finally, a \textbf{Deep Ensemble} composed of five MLPs following \protect\hyperlink{ref-lakshminarayanan2016simple}{{[}30{]}} that serves as our only probabilistic classifier. We have chosen to work with deep ensembles both for their simplicity and effectiveness at modelling predictive uncertainty. They are also the model of choice in \protect\hyperlink{ref-schut2021generating}{{[}6{]}}. The network architectures are kept simple (top half of Table \ref{tab:architecture}), since we are only marginally concerned with achieving good initial classifier performance.

The Latent Space generator relies on a separate generative model. Following the authors of both REVISE and CLUE we use Variational Autoencoders (\textbf{VAE}) for this purpose. As with the classifiers, we deliberately choose to work with fairly simple architectures (bottom half of Table \ref{tab:architecture}). More expressive generative models generally also lead to more meaningful counterfactuals produced by Latent Space generators. But in our view, this should simply be considered as a vulnerability of counterfactual generators that rely on surrogate models to learn realistic representations of the underlying data.

\begin{table}

\caption{\label{tab:architecture}Neural network architectures and training parameters.}
\centering
\resizebox{\linewidth}{!}{
\begin{tabular}[t]{llrlrllr}
\toprule
 & Data & Hidden Dim. & Latent Dim. & Hidden Layers & Batch & Dropout & Epochs\\
\midrule
\addlinespace[0.3em]
\multicolumn{8}{l}{\textbf{MLP}}\\
\hspace{1em} & Synthetic & 32 & - & 1 & - & - & 100\\
\cmidrule{2-8}
\hspace{1em} & Real-World & 64 & - & 2 & 500 & 0.1 & 100\\
\cmidrule{1-8}
\addlinespace[0.3em]
\multicolumn{8}{l}{\textbf{VAE}}\\
\hspace{1em} & Synthetic & 32 & 2 & 1 & - & - & 100\\
\cmidrule{2-8}
\hspace{1em} & Real-World & 32 & 8 & 1 & - & - & 250\\
\bottomrule
\end{tabular}}
\end{table}

\hypertarget{empirical-data}{%
\subsection{\texorpdfstring{\(\mathcal{D}\)---Data}{\textbackslash mathcal\{D\}---Data}}\label{empirical-data}}

We have chosen to work with both synthetic and real-world datasets. Using synthetic data allows us to impose distributional properties that may affect the resulting recourse dynamics. Following \protect\hyperlink{ref-upadhyay2021robust}{{[}14{]}}, we generate synthetic data in \(\mathbb{R}^2\) to also allow for a visual interpretation of the results. Real-world data is used in order to assess if endogenous dynamics also occur in higher-dimensional settings.

\hypertarget{synthetic-data}{%
\subsubsection{Synthetic data}\label{synthetic-data}}

\begin{figure}

{\centering \includegraphics[width=0.9\linewidth]{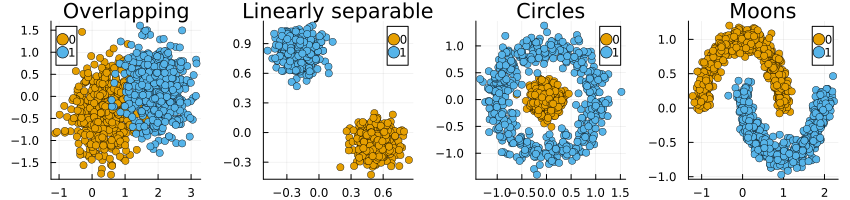} 

}

\caption{Synthetic classification datasets used in our experiments. Samples from the negative class ($y=0$) are marked in blue while samples of the positive class ($y=1$) are marked in orange.}\label{fig:synthetic-data}
\end{figure}

We use four synthetic binary classification datasets consisting of 1000 samples each: \textbf{Overlapping}, \textbf{Linearly Separable}, \textbf{Circles} and \textbf{Moons} (Figure \ref{fig:synthetic-data}).

Ex-ante we expect to see that by construction, Wachter will create a new cluster of counterfactual instances in the proximity of the initial decision boundary as we saw in Figure \ref{fig:poc}. Thus, the choice of a black-box model may have an impact on the counterfactual paths. For generators that use latent space search (REVISE \protect\hyperlink{ref-joshi2019realistic}{{[}7{]}}, CLUE \protect\hyperlink{ref-antoran2020getting}{{[}9{]}}) or rely on (and have access to) probabilistic models (CLUE \protect\hyperlink{ref-antoran2020getting}{{[}9{]}}, Greedy \protect\hyperlink{ref-schut2021generating}{{[}6{]}}) we expect that counterfactuals will end up in regions of the target domain that are densely populated by training samples. Of course, this expectation hinges on how effective said probabilistic models are at capturing predictive uncertainty. Finally, we expect to see the counterfactuals generated by DiCE to be diversely spread around the feature space inside the target class\footnote{As we mentioned earlier, the diversity constraint used by DiCE is only effective when at least two counterfactuals are being generated. We have therefore decided to always generate 5 counterfactuals for each generator and randomly pick one of them.}. In summary, we expect that the endogenous shifts induced by Wachter outsize those of all other generators since Wachter is not explicitly concerned with generating what we have defined as meaningful counterfactuals.

\hypertarget{real-world-data}{%
\subsubsection{Real-world data}\label{real-world-data}}

We use three different real-world datasets from the Finance and Economics domain, all of which are tabular and can be used for binary classification. Firstly, we use the \textbf{Give Me Some Credit} dataset which was open-sourced on Kaggle for the task to predict whether a borrower is likely to experience financial difficulties in the next two years \protect\hyperlink{ref-kaggle2011give}{{[}31{]}}, originally consisting of 250,000 instances with 11 numerical attributes. Secondly, we use the \textbf{UCI defaultCredit} dataset \protect\hyperlink{ref-yeh2009comparisons}{{[}32{]}}, a benchmark dataset that can be used to train binary classifiers to predict the binary outcome variable of whether credit card clients default on their payment. In its raw form, it consists of 23 explanatory variables: 4 categorical features relating to demographic attributes and 19 continuous features largely relating to individuals' payment histories and amount of credit outstanding. Both datasets have been used in the literature on AR before (see for example \protect\hyperlink{ref-pawelczyk2021carla}{{[}3{]}}, \protect\hyperlink{ref-joshi2019realistic}{{[}7{]}} and \protect\hyperlink{ref-ustun2019actionable}{{[}12{]}}), presumably because they constitute real-world classification tasks involving individuals that compete for access to credit.

As a third dataset, we include the \textbf{California Housing} dataset derived from the 1990 U.S. census and sourced through scikit-learn \protect\hyperlink{ref-pace1997sparse}{{[}34{]}}. It consists of 8 continuous features that can be used to predict the median house price for California districts. The continuous outcome variable is binarized as \(\tilde{y}=\mathbb{I}_{y>\text{median}(Y)}\) indicating whether or not the median house price of a given district is above the median of all districts. While we have not seen this dataset used in the previous literature on AR, others have used the Boston Housing dataset in a similar fashion \protect\hyperlink{ref-schut2021generating}{{[}6{]}}. We initially also conducted experiments on that dataset, but eventually discarded it due to surrounding ethical concerns \protect\hyperlink{ref-carlisle2019racist}{{[}35{]}}.

Since the simulations involve generating counterfactuals for a significant proportion of the entire sample of individuals, we have randomly undersampled each dataset to yield balanced subsamples consisting of 5,000 individuals each. We have also standardized all continuous explanatory features since our chosen classifiers are sensitive to scale.

\hypertarget{ggenerators}{%
\subsection{\texorpdfstring{\(G\)---Generators}{G---Generators}}\label{ggenerators}}

All generators introduced earlier are included in the experiments: Wachter \protect\hyperlink{ref-wachter2017counterfactual}{{[}4{]}}, REVISE \protect\hyperlink{ref-joshi2019realistic}{{[}7{]}}, CLUE \protect\hyperlink{ref-antoran2020getting}{{[}9{]}}, DiCE \protect\hyperlink{ref-mothilal2020explaining}{{[}8{]}} and Greedy \protect\hyperlink{ref-schut2021generating}{{[}6{]}}. In addition, we introduce two new generators in Section \ref{mitigate} that directly address the issue of endogenous domain and model shifts. We also test to what extent it may be beneficial to combine ideas underlying the various generators.

\hypertarget{empirical-2}{%
\section{Experiments}\label{empirical-2}}

Below, we first present our main experimental findings regarding these questions. We conclude this section with a brief recap providing answers to all of these questions.

\hypertarget{endogenous-macrodynamics}{%
\subsection{Endogenous Macrodynamics}\label{endogenous-macrodynamics}}

We start this section off with the key high-level observations. Across all datasets (synthetic and real), classifiers and counterfactual generators we observe either most or all of the following dynamics at varying degrees:

\begin{itemize}
\tightlist
\item
  Statistically significant domain and model shifts as measured by MMD.
\item
  A deterioration in out-of-sample model performance as measured by the F-Score evaluated on a test sample. In many cases this drop in performance is substantial.
\item
  Significant perturbations to the model parameters as well as an increase in the model's decisiveness.
\item
  Disagreement between the original and retrained model, in some cases large.
\end{itemize}

There is also some clear heterogeneity across the results:

\begin{itemize}
\tightlist
\item
  The observed dynamics are generally of the highest magnitude for the linear classifier. Differences in results for the MLP and Deep Ensemble are mostly negligible.
\item
  The reduction in model performance appears to be most severe when classes are not perfectly separable or the initial model performance was weak, to begin with.
\item
  Except for the Greedy generator, all other generators generally perform somewhat better overall than the baseline (Wachter) as expected.
\end{itemize}

\begin{figure}

{\centering \includegraphics[width=0.9\linewidth]{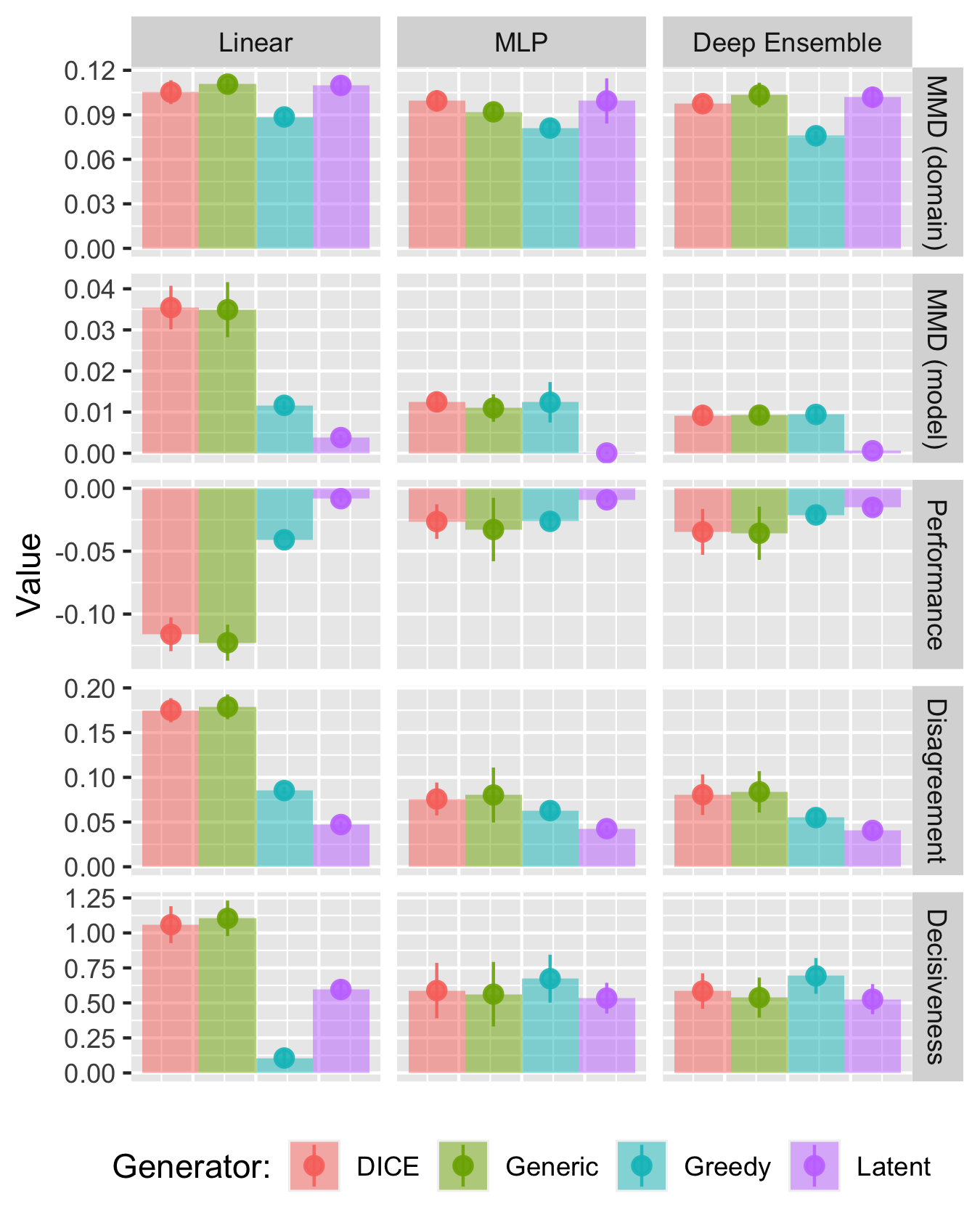} 

}

\caption{Results for synthetic data with overlapping classes. The shown model MMD (PP MMD) was computed over a mesh grid of 1,000 points. Error bars indicate the standard deviation across folds.}\label{fig:syn}
\end{figure}

Focusing first on synthetic data, Figure \ref{fig:syn} presents our findings for the dataset with overlapping classes. It shows the resulting values for some of our evaluation metrics at the end of the experiment, after all \(T=50\) rounds, along with error bars indicating the variation across folds.

The top row shows the estimated domain shifts. While it is difficult to interpret the exact magnitude of MMD, we can see that the values are different from zero and there is essentially no variation across our five folds. For the domain shifts, the Greedy generator induces the smallest shifts. In general, we have observed the opposite.

The second row shows the estimated model shifts, where here we have used the grid approach explained earlier. As with the domain shifts, the observed values are clearly different from zero and variation across folds is once again small. In this case, the results for this particular dataset very much reflect the broader patterns we have observed: Latent Space (LS) generators induce the smallest shifts, followed by DiCE, then Wachter and finally Greedy.

The same broad pattern also emerges in the third row: we observe the smallest deterioration in model performance for LS generators, albeit we still find a reduction in the F-Score of around 5-10 percentage points on average. Related to this, the bottom two rows indicate that the retrained classifiers disagree with their initial counterparts on the classification of up to nearly 25 per cent of the individuals. We also note that the final classifiers are more decisive, although as we noted earlier this may to some extent just be a byproduct of retraining the model throughout the experiment.

Figure \ref{fig:syn} also indicates that the estimated effects are strongest for the simplest linear classifier, a pattern that we have observed fairly consistently. Conversely, there is virtually no difference in outcomes between the deep ensemble and the MLP. It is possible that the deep ensembles simply fail to capture predictive uncertainty well and hence counterfactual generators like Greedy, which explicitly addresses this quantity, fail to work as expected.

The findings for the other synthetic datasets are broadly consistent with the observations above. For the Moons data, the same broad patterns emerge, although in this case, the Greedy generator induces comparably strong shifts in some cases. For the Circles data, model shifts and performance deterioration are quantitatively much smaller than what we can observe in Figure \ref{fig:syn} and in many cases insignificant. For the Linearly Separable data we also find substantial domain and model shifts, but almost no reduction in model performance.\footnote{You can find a granular overview of all results including bootstraps in our online companion: \url{https://www.paltmeyer.com/endogenous-macrodynamics-in-algorithmic-recourse/}.}

Finally, it is also worth noting that the observed dynamics and patterns are consistent throughout the experiment. That is to say that we start observing shifts already after just a few rounds and these tend to increase proportionately for the different generators over the course of the experiment.

\begin{figure}

{\centering \includegraphics[width=0.9\linewidth]{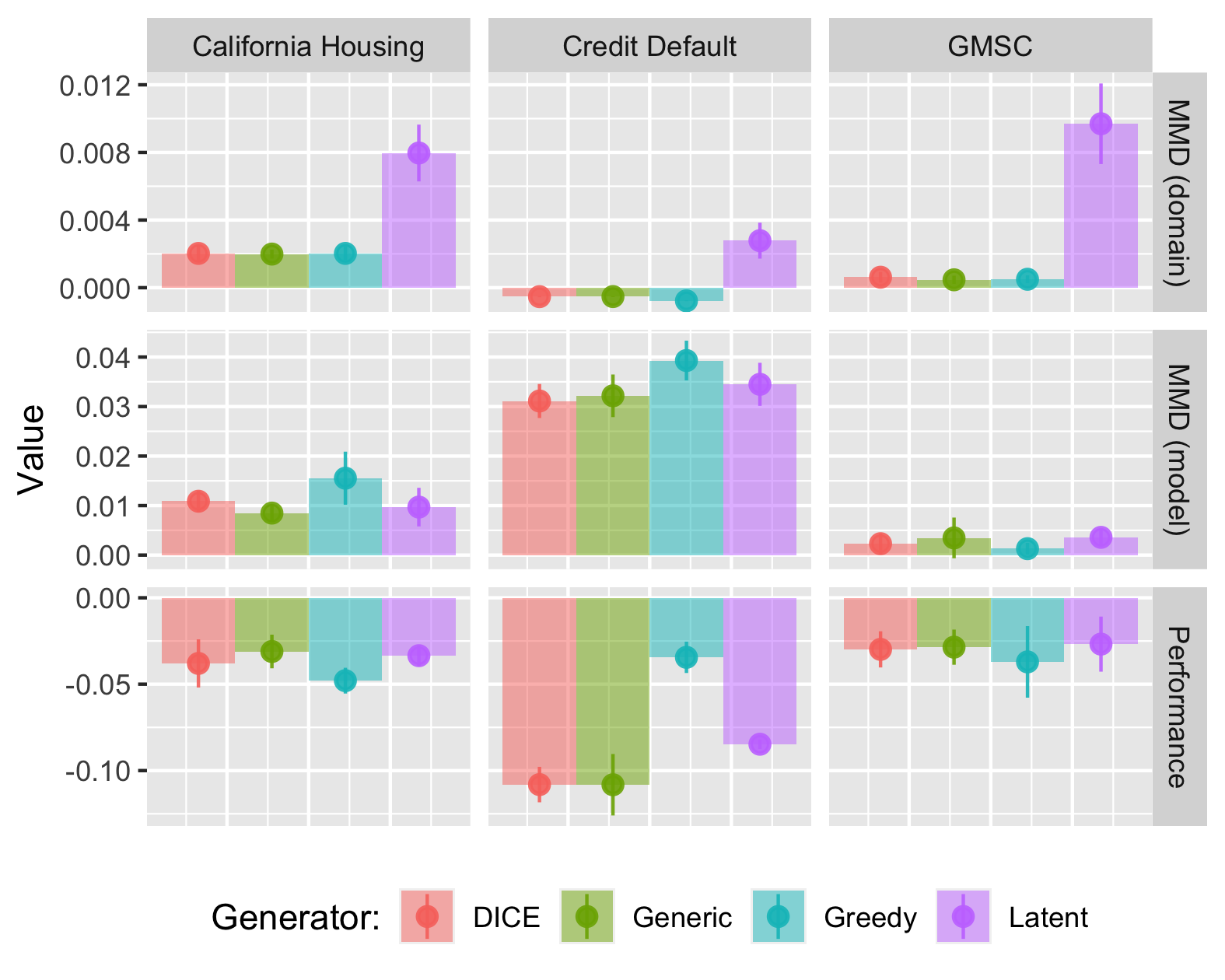} 

}

\caption{Results for deep ensemble using real-world datasets. The shown model MMD (PP MMD) was computed over actual samples, rather than a mesh grid. Error bars indicate the standard deviation across folds.}\label{fig:real}
\end{figure}

Turning to the real-world data we will go through the findings presented in Figure \ref{fig:real}, where each column corresponds to one of the three data sets. The results shown here are for the deep ensemble, which once again largely resemble those for the MLP. Starting from the top row, we find significant domain shifts of varying magnitudes. Latent Space search induces shifts that are orders of magnitude higher than for the other generators, which generally induce significant but small shifts.

Model shifts are shown in the middle row of Figure \ref{fig:real}: the estimated PP MMD is statistically significant across the board and in some cases much larger than in others. We find no evidence that LS search helps to mitigate model shifts, as we did before for the synthetic data. Since these real-world datasets are arguably more complex than the synthetic data, the generative model can be expected to have a harder time learning the data-generating process and hence this increased difficulty appears to affect the performance of REVISE/CLUE.

The out-of-sample model performance also deteriorates across the board and substantially so: the largest average reduction in F-Scores of more than 10 percentage points is observed for the Credit Default dataset. For this dataset we achieved the lowest initial model performance, indicating once again that weaker classifiers may be more exposed to endogenous dynamics. As with the synthetic data, the estimates for logistic regression are qualitatively in line with the above, but quantitatively even more pronounced.

To recap, we answer our research questions: firstly, endogenous dynamics do emerge in our experiments (RQ \ref{prp:shifts}) and we find them substantial enough to be considered costly (RQ \ref{prp:costs}); secondly, the choice of the counterfactual generator matters, with Latent Space search generally having a dampening effect (RQ \ref{prp:het}). The observed dynamics, therefore, seem to be driven by a discrepancy between counterfactual outcomes that minimize costs to individuals and outcomes that comply with the data-generating process (RQ \ref{prp:drive}).

\hypertarget{mitigate}{%
\section{Mitigation Strategies and Experiments}\label{mitigate}}

Having established in the previous section that endogenous macrodynamics in AR are substantial enough to warrant our attention, in this section we ask ourselves:

\begin{proposition}[Mitigation Strategies]
\protect\hypertarget{prp:mitigate}{}\label{prp:mitigate}What are potential mitigation strategies with respect to endogenous macrodynamics in AR?
\end{proposition}

We propose and test several simple mitigation strategies. All of them essentially boil down to one simple principle: to avoid domain and model shifts, the generated counterfactuals should comply as much as possible with the true data-generating process. This principle is really at the core of Latent Space (LS) generators, and hence it is not surprising that we have found these types of generators to perform comparably well in the previous section. But as we have mentioned earlier, generators that rely on separate generative models carry an additional computational burden and, perhaps more importantly, their performance hinges on the performance of said generative models. Fortunately, it turns out that we can use a number of other, much simpler strategies.

\hypertarget{more-conservative-decision-thresholds}{%
\subsection{More Conservative Decision Thresholds}\label{more-conservative-decision-thresholds}}

The most obvious and trivial mitigation strategy is to simply choose a higher decision threshold \(\gamma\). This threshold determines when a counterfactual should be considered valid. Under \(\gamma=0.5\), counterfactuals will end up near the decision boundary by construction. Since this is the region of maximal aleatoric uncertainty, the classifier is bound to be thrown off. By setting a more conservative threshold, we can avoid this issue to some extent. A drawback of this approach is that a classifier with high decisiveness may classify samples with high confidence even far away from the training data.

\hypertarget{classifier-preserving-roar-claproar}{%
\subsection{Classifier Preserving ROAR (ClaPROAR)}\label{classifier-preserving-roar-claproar}}

Another strategy draws inspiration from ROAR \protect\hyperlink{ref-upadhyay2021robust}{{[}14{]}}: to preserve the classifier, we propose to explicitly penalize the loss it incurs when evaluated on the counterfactual \(x^\prime\) at given parameter values. Recall that \(\text{extcost}(\cdot)\) denotes what we had defined as the external cost in Equation \eqref{eq:collective}. Formally, we let

\begin{equation}
\begin{aligned}
\text{extcost}(f(\mathbf{s}^\prime)) = l(M(f(\mathbf{s}^\prime)),y^\prime) \label{eq:clap}
\end{aligned}
\end{equation}

for each counterfactual \(k\) where \(l\) denotes the loss function used to train \(M\). This approach, which we refer to as \textbf{ClaPROAR}, is based on the intuition that (endogenous) model shifts will be triggered by counterfactuals that increase classifier loss. It is closely linked to the idea of choosing a higher decision threshold, but is likely better at avoiding the potential pitfalls associated with highly decisive classifiers. It also makes the private vs.~external cost trade-off more explicit and hence manageable.

\hypertarget{gravitational-counterfactual-explanations}{%
\subsection{Gravitational Counterfactual Explanations}\label{gravitational-counterfactual-explanations}}

Yet another strategy extends Wachter as follows: instead of only penalizing the distance of the individuals' counterfactual to its factual, we propose penalizing its distance to some sensible point in the target domain, for example, the subsample average \(\bar{x}^*=\text{mean}(x)\), \(x \in \mathcal{D}_1\):

\begin{equation}
\begin{aligned}
\text{extcost}(f(\mathbf{s}^\prime)) = \text{dist}(f(\mathbf{s}^\prime),\bar{x}^*)  \label{eq:grav}
\end{aligned}
\end{equation}

Once again we can put this in the context of Equation \eqref{eq:collective}: the former penalty can be thought of here as the private cost incurred by the individual, while the latter reflects the external cost incurred by other individuals. Higher choices of \(\lambda_2\) relative to \(\lambda_1\) will lead counterfactuals to gravitate towards the specified point \(\bar{x}^*\) in the target domain. In the remainder of this paper, we will therefore refer to this approach as \textbf{Gravitational} generator, when we investigate its usefulness for mitigating endogenous macrodynamics\footnote{Note that despite the naming conventions, our goal here is not to provide yet more counterfactual generators. Rather than looking at them as isolated entities, we believe and demonstrate that different approaches can be effectively combined.}.

\begin{figure}

{\centering \includegraphics[width=0.9\linewidth]{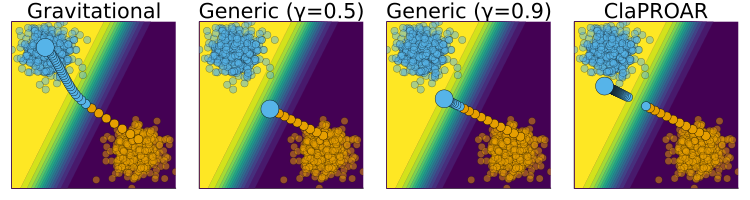} 

}

\caption{Illustrative example demonstrating the properties of the various mitigation strategies. Samples from the negative class ($y=0$) are marked in orange while samples of the positive class ($y=1$) are marked in blue.}\label{fig:mitigation}
\end{figure}

Figure \ref{fig:mitigation} shows an illustrative example that demonstrates the differences in counterfactual outcomes when using the various mitigation strategies compared to the baseline approach, that is, Wachter with \(\gamma=0.5\): choosing a higher decision threshold pushes the counterfactual a little further into the target domain; this effect is even stronger for ClaPROAR; finally, using the Gravitational generator the counterfactual ends up all the way inside the target domain in the neighbourhood of \(\bar{x}^*\)\footnote{In order for the Gravitational generator and ClaPROAR to work as expected, one needs to ensure that counterfactual search continues, independent of the threshold probability \(\gamma\).}. Linking these ideas back to Example \ref{exm:student}, the mitigation strategies help ensure that the recommended recourse actions are substantial enough to truly lead to an increase in the probability that the admitted student eventually graduates.

\begin{figure}

{\centering \includegraphics[width=0.9\linewidth]{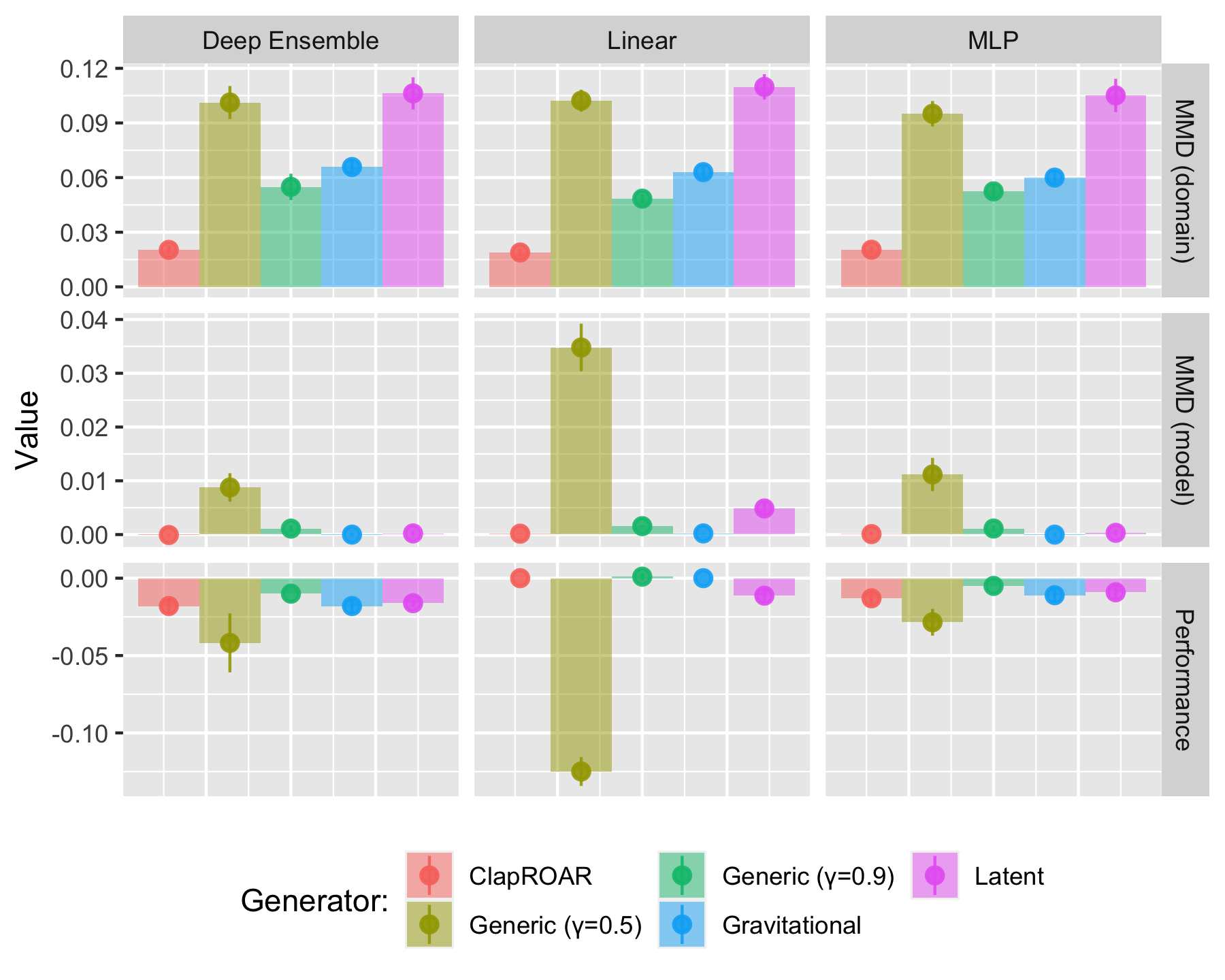} 

}

\caption{The differences in counterfactual outcomes when using the various mitigation strategies compared to the baseline approach, that is Wachter with $\gamma=0.5$. Results for synthetic data with overlapping classes. The shown model MMD (PP MMD) was computed over a mesh grid of points. Error bars indicate the standard deviation across folds.}\label{fig:mitigate-results}
\end{figure}

Our findings indicate that all three mitigation strategies are at least at par with LS generators with respect to their effectiveness at mitigating domain and model shifts. Figure \ref{fig:mitigate-results} presents a subset of the evaluation metrics for our synthetic data with overlapping classes. The top row in Figure \ref{fig:mitigate-results} indicates that while domain shifts are of roughly the same magnitude for both Wachter and LS generators, our proposed strategies effectively mitigate these shifts. ClaPROAR appears to be particularly effective, which is positively surprising since it is designed to explicitly address model shifts, not domain shifts. As evident from the middle row in Figure \ref{fig:mitigate-results} model shifts can also be reduced: for the deep ensemble LS search yields results that are at par with the mitigation strategies, while for both the simple MLP and logistic regression our simple strategies are more effective. The same overall pattern can be observed for out-of-sample model performance. Concerning the other synthetic datasets, for the Moons dataset, the emerging patterns are largely the same, but the estimated model shifts are insignificant as noted earlier; the same holds for the Circles dataset, but there is no significant reduction in model performance for our neural networks; in the case of linearly separable data, we find the Gravitational generator to be most effective at mitigating shifts.

\begin{figure}

{\centering \includegraphics[width=0.9\linewidth]{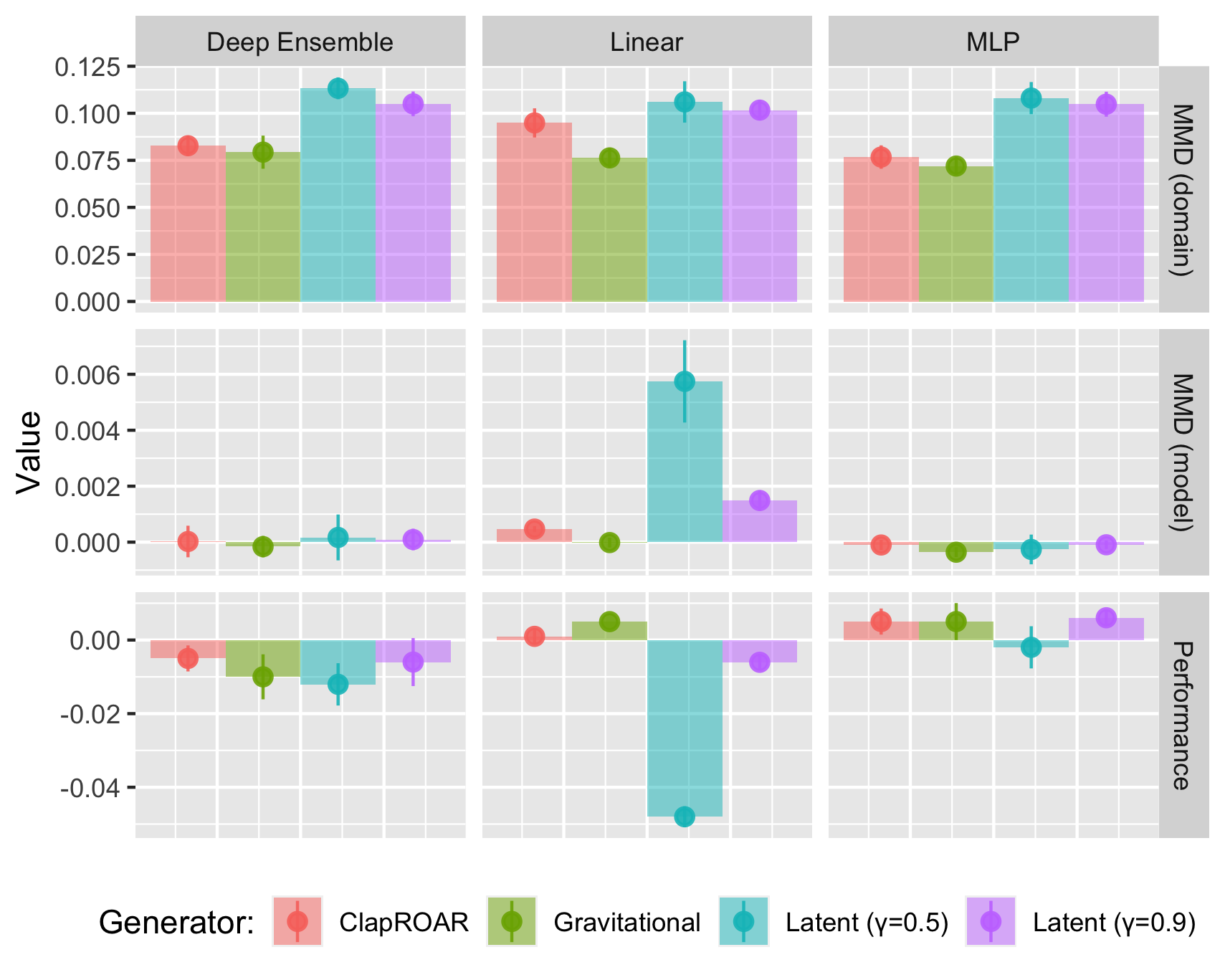} 

}

\caption{Combining various mitigation strategies with LS search. Results for synthetic data with overlapping classes. The shown model MMD (PP MMD) was computed over a mesh grid of points. Error bars indicate the standard deviation across folds.}\label{fig:mitigate-latent-results}
\end{figure}

An interesting finding is also that the proposed strategies have a complementary effect when used in combination with LS generators. In experiments we conducted on the synthetic data, the benefits of LS generators were exacerbated further when using a more conservative threshold or combining it with the penalties underlying Gravitational and ClaPROAR. In Figure \ref{fig:mitigate-latent-results} the conventional LS generator with \(\gamma=0.5\) serves as our baseline. Evidently, being more conservative or using one of our proposed penalties decreases the estimated domain and model shifts, in some cases beyond significance.

\begin{figure}

{\centering \includegraphics[width=0.9\linewidth]{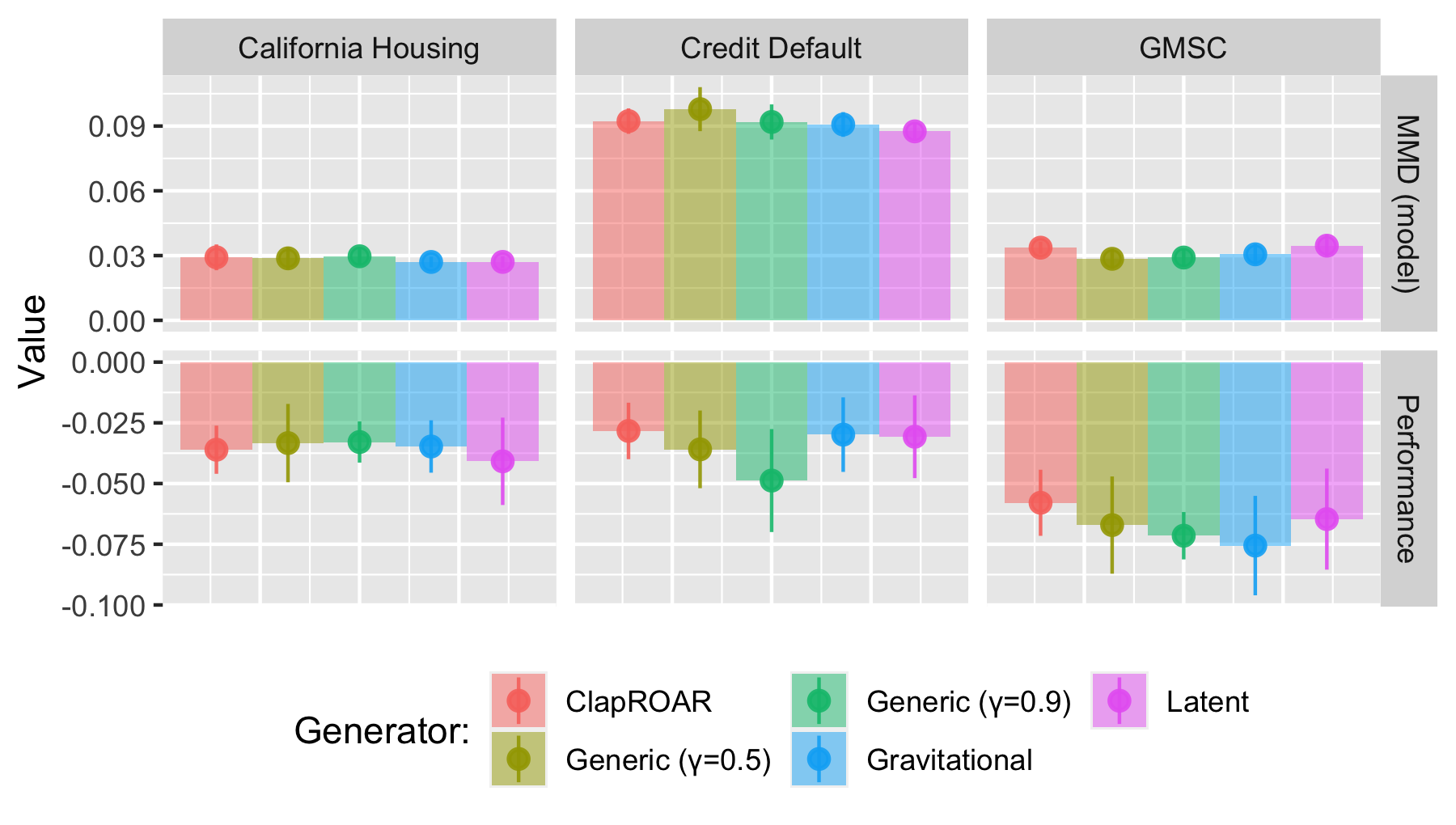} 

}

\caption{The differences in counterfactual outcomes when using the various mitigation strategies compared to the baseline approach, that is Wachter with $\gamma=0.5$. Results for the MLP using real-world datasets. The shown model MMD (PP MMD) was computed over actual samples, rather than a mesh grid. Error bars indicate the standard deviation across folds.}\label{fig:mitigate-real-world-results}
\end{figure}

Finally, Figure \ref{fig:mitigate-real-world-results} shows the results for our real-world data. We note that for both the California Housing and GMSC data, ClaPROAR does have an attenuating effect on model performance deterioration\footnote{Estimated domain shifts (not shown) were largely insubstantial, as in Figure \ref{fig:real} in the previous section.}. Overall, the results are less significant, possibly because a somewhat smaller share of individuals from the non-target group received recourse than in the synthetic case\footnote{In \href{https://github.com/pat-alt/endogenous-macrodynamics-in-algorithmic-recourse/releases/tag/dec-2022}{earlier experiments} we moved a larger share of individuals and the results more clearly favoured our mitigation strategies.}.

\hypertarget{discussion}{%
\section{Discussion}\label{discussion}}

Our results in Section \ref{empirical-2} indicate that state-of-the-art approaches to Algorithmic Recourse induce substantial domain and model shift if implemented at scale in practice. These induced shifts can and should be considered as an (expected) external cost of individual recourse. While they do not affect the individual directly as long as we look at the individual in isolation, they can be seen to affect the broader group of stakeholders in automated data-driven decision-making. We have seen, for example, that out-of-sample model performance generally deteriorates in our simulation experiments. In practice, this can be seen as a cost to model owners, that is the group of stakeholders using the model as a decision-making tool. As we have set out in Example \ref{exm:student} of our introduction, these model owners may be unwilling to carry that cost, and hence can be expected to stop offering recourse to individuals altogether. This in turn is costly to those individuals that would otherwise derive utility from being offered recourse.

So, where does this leave us? We would argue that the expected external costs of individual recourse should be shared by all stakeholders. The most straightforward way to achieve this is to introduce a penalty for external costs in the counterfactual search objective function, as we have set out in Equation \eqref{eq:collective}. This will on average lead to more costly counterfactual outcomes, but may help to avoid extreme scenarios, in which minimal-cost recourse is reserved to a tiny minority of individuals. We have shown various types of shift-mitigating strategies that can be used to this end. Since all of these strategies can be seen simply as a specific adaption of Equation \eqref{eq:collective}, they can be applied to any of the various counterfactual generators studied here.

\hypertarget{limit}{%
\section{Limitations and Future Work}\label{limit}}

While we believe that this work constitutes a valuable starting point for addressing existing issues in Algorithmic Recourse from a fresh perspective, we are aware of several of its limitations. In the following, we highlight some of these and point to avenues for future research.

\hypertarget{private-vs.-external-costs}{%
\subsection{Private vs.~External Costs}\label{private-vs.-external-costs}}

Perhaps the most crucial shortcoming of our work is that we merely point out that there exists a trade-off between private costs to the individual and external costs to the collective of stakeholders. We fall short of providing any definitive answers as to how that trade-off may be resolved in practice. The mitigation strategies we have proposed here provide a good starting point, but they are ad-hoc extensions of the existing AR framework. An interesting idea to explore in future work could be the potential for Pareto optimal Algorithmic Recourse, that is, a collective recourse outcome in which no single individual can be made better off, without making at least one other individual worse off. This type of work would be interdisciplinary and could help to formalize some of the concepts presented in this work.

\hypertarget{experimental-setup}{%
\subsection{Experimental Setup}\label{experimental-setup}}

The experimental setup proposed here is designed to mimic a real-world recourse process in a simple fashion. In practice, models are updated regularly \protect\hyperlink{ref-upadhyay2021robust}{{[}14{]}}. We also find it plausible to assume that the implementation of recourse happens periodically for different individuals, rather than all at once at time \(t=0\). That being said, our experimental design is a vast over-simplification of potential real-world scenarios. In practice, any endogenous shifts that may occur can be expected to be entangled with exogenous shifts of the nature investigated in Upadhyay et al. \protect\hyperlink{ref-upadhyay2021robust}{{[}14{]}}. We also make implicit assumptions about the utility functions of the involved agents that may well be too simple: individuals seeking recourse are assumed to always implement the proposed Counterfactual Explanations; conversely, the agent in charge of the model \(M\) is assumed to always treat individuals that have implemented valid recourse as if they were truly now in the target class.

\hypertarget{causal-modelling}{%
\subsection{Causal Modelling}\label{causal-modelling}}

In this work, we have focused on popular counterfactual generators that do not incorporate any causal knowledge. The generated perturbations therefore may involve changes to variables that affect the outcome predicted by the black-box model, but not the true outcome. The implementation of such changes is typically described as \textbf{gaming} \protect\hyperlink{ref-miller2020strategic}{{[}36{]}}, although they need not be driven by adversarial intentions: in Example \ref{exm:student}, student applicants may dutifully focus on acquiring credentials that help them to be admitted to university, but ultimately not to improve their chances of success at completing their degree \protect\hyperlink{ref-barocas2022fairness}{{[}37{]}}. Preventing such actions may help to avoid the dynamics we have pointed to in this work. Future work would likely benefit from including recent approaches to AR that incorporate causal knowledge such as Karimi et al. \protect\hyperlink{ref-karimi2021algorithmic}{{[}13{]}}.

\hypertarget{classifiers}{%
\subsection{Classifiers}\label{classifiers}}

For reasons stated earlier, we have limited our analysis to differentiable linear and non-linear classifiers, in particular logistic regression and deep neural networks. While these sorts of classifiers have also typically been analyzed in the existing literature on Counterfactual Explanations and Algorithmic Recourse, they represent only a subset of popular machine learning models employed in practice. Despite the success and popularity of deep learning in the context of high-dimensional data such as image, audio and video, empirical evidence suggests that other models such as boosted decision trees may have an edge when it comes to lower-dimensional tabular datasets, such as the ones considered here (\protect\hyperlink{ref-borisov2021deep}{{[}38{]}}, \protect\hyperlink{ref-grinsztajn2022why}{{[}39{]}}).

\hypertarget{limit-data}{%
\subsection{Data}\label{limit-data}}

Largely in line with the existing literature on Algorithmic Recourse, we have limited our analysis of real-world data to three commonly used benchmark datasets that involve binary prediction tasks. Future work may benefit from including novel datasets or extending the analysis to multi-class or regression problems, the latter arguably representing the most common objective in Finance and Economics.

\hypertarget{conclusion}{%
\section{Concluding Remarks}\label{conclusion}}

This work has revisited and extended some of the most general and defining concepts underlying the literature on Counterfactual Explanations and, in particular, Algorithmic Recourse. We demonstrate that long-held beliefs as to what defines optimality in AR, may not always be suitable. Specifically, we run experiments that simulate the application of recourse in practice using various state-of-the-art counterfactual generators and find that all of them induce substantial domain and model shifts. We argue that these shifts should be considered as an expected external cost of individual recourse and call for a paradigm shift from individual to collective recourse in these types of situations. By proposing an adapted counterfactual search objective that incorporates this cost, we make that paradigm shift explicit. We show that this modified objective lends itself to mitigation strategies that can be used to effectively decrease the magnitude of induced domain and model shifts. Through our work, we hope to inspire future research on this important topic. To this end we have open-sourced all of our code along with a Julia package: \href{https://anonymous.4open.science/r/AlgorithmicRecourseDynamics/README.md}{\texttt{AlgorithmicRecourseDynamics.jl}}. Future researchers should find it easy to replicate, modify and extend the simulation experiments presented here and apply them to their own custom counterfactual generators.

\hypertarget{acknowledgements}{%
\section*{Acknowledgements}\label{acknowledgements}}
\addcontentsline{toc}{section}{Acknowledgements}

Some of the members of TU Delft were partially funded by ICAI AI for Fintech Research, an ING --- TU Delft collaboration.

\hypertarget{references}{%
\section*{References}\label{references}}
\addcontentsline{toc}{section}{References}

\hypertarget{refs}{}
\begin{CSLReferences}{0}{0}
\leavevmode\vadjust pre{\hypertarget{ref-oneil2016weapons}{}}%
\CSLLeftMargin{{[}1{]} }%
\CSLRightInline{C. O'Neil, \emph{Weapons of math destruction: {How} big data increases inequality and threatens democracy}. {Crown}, 2016.}

\leavevmode\vadjust pre{\hypertarget{ref-rudin2019stop}{}}%
\CSLLeftMargin{{[}2{]} }%
\CSLRightInline{C. Rudin, {``Stop explaining black box machine learning models for high stakes decisions and use interpretable models instead,''} \emph{Nature Machine Intelligence}, vol. 1, no. 5, pp. 206--215, 2019.}

\leavevmode\vadjust pre{\hypertarget{ref-pawelczyk2021carla}{}}%
\CSLLeftMargin{{[}3{]} }%
\CSLRightInline{M. Pawelczyk, S. Bielawski, J. van den Heuvel, T. Richter, and G. Kasneci, {``Carla: A python library to benchmark algorithmic recourse and counterfactual explanation algorithms,''} 2021. Available: \url{https://arxiv.org/abs/2108.00783}}

\leavevmode\vadjust pre{\hypertarget{ref-wachter2017counterfactual}{}}%
\CSLLeftMargin{{[}4{]} }%
\CSLRightInline{S. Wachter, B. Mittelstadt, and C. Russell, {``Counterfactual explanations without opening the black box: {Automated} decisions and the {GDPR},''} \emph{Harv. JL \& Tech.}, vol. 31, p. 841, 2017.}

\leavevmode\vadjust pre{\hypertarget{ref-altmeyer2022counterfactualexplanations}{}}%
\CSLLeftMargin{{[}5{]} }%
\CSLRightInline{P. Altmeyer, {``{CounterfactualExplanations}.jl - a {Julia} package for {Counterfactual Explanations} and {Algorithmic Recourse}.''} 2022. Available: \url{https://github.com/pat-alt/CounterfactualExplanations.jl}}

\leavevmode\vadjust pre{\hypertarget{ref-schut2021generating}{}}%
\CSLLeftMargin{{[}6{]} }%
\CSLRightInline{L. Schut \emph{et al.}, {``Generating {Interpretable Counterfactual Explanations By Implicit Minimisation} of {Epistemic} and {Aleatoric Uncertainties},''} in \emph{International {Conference} on {Artificial Intelligence} and {Statistics}}, 2021, pp. 1756--1764.}

\leavevmode\vadjust pre{\hypertarget{ref-joshi2019realistic}{}}%
\CSLLeftMargin{{[}7{]} }%
\CSLRightInline{S. Joshi, O. Koyejo, W. Vijitbenjaronk, B. Kim, and J. Ghosh, {``Towards realistic individual recourse and actionable explanations in black-box decision making systems,''} 2019. Available: \url{https://arxiv.org/abs/1907.09615}}

\leavevmode\vadjust pre{\hypertarget{ref-mothilal2020explaining}{}}%
\CSLLeftMargin{{[}8{]} }%
\CSLRightInline{R. K. Mothilal, A. Sharma, and C. Tan, {``Explaining machine learning classifiers through diverse counterfactual explanations,''} in \emph{Proceedings of the 2020 {Conference} on {Fairness}, {Accountability}, and {Transparency}}, 2020, pp. 607--617.}

\leavevmode\vadjust pre{\hypertarget{ref-antoran2020getting}{}}%
\CSLLeftMargin{{[}9{]} }%
\CSLRightInline{J. Antorán, U. Bhatt, T. Adel, A. Weller, and J. M. Hernández-Lobato, {``Getting a clue: {A} method for explaining uncertainty estimates,''} 2020. Available: \url{https://arxiv.org/abs/2006.06848}}

\leavevmode\vadjust pre{\hypertarget{ref-karimi2020survey}{}}%
\CSLLeftMargin{{[}10{]} }%
\CSLRightInline{A.-H. Karimi, G. Barthe, B. Schölkopf, and I. Valera, {``A survey of algorithmic recourse: Definitions, formulations, solutions, and prospects,''} 2020. Available: \url{https://arxiv.org/abs/2010.04050}}

\leavevmode\vadjust pre{\hypertarget{ref-verma2020counterfactual}{}}%
\CSLLeftMargin{{[}11{]} }%
\CSLRightInline{S. Verma, J. Dickerson, and K. Hines, {``Counterfactual explanations for machine learning: {A} review,''} 2020. Available: \url{https://arxiv.org/abs/2010.10596}}

\leavevmode\vadjust pre{\hypertarget{ref-ustun2019actionable}{}}%
\CSLLeftMargin{{[}12{]} }%
\CSLRightInline{B. Ustun, A. Spangher, and Y. Liu, {``Actionable recourse in linear classification,''} in \emph{Proceedings of the {Conference} on {Fairness}, {Accountability}, and {Transparency}}, 2019, pp. 10--19.}

\leavevmode\vadjust pre{\hypertarget{ref-karimi2021algorithmic}{}}%
\CSLLeftMargin{{[}13{]} }%
\CSLRightInline{A.-H. Karimi, B. Schölkopf, and I. Valera, {``Algorithmic recourse: From counterfactual explanations to interventions,''} in \emph{Proceedings of the 2021 {ACM Conference} on {Fairness}, {Accountability}, and {Transparency}}, 2021, pp. 353--362.}

\leavevmode\vadjust pre{\hypertarget{ref-upadhyay2021robust}{}}%
\CSLLeftMargin{{[}14{]} }%
\CSLRightInline{S. Upadhyay, S. Joshi, and H. Lakkaraju, {``Towards {Robust} and {Reliable Algorithmic Recourse},''} 2021. Available: \url{https://arxiv.org/abs/2102.13620}}

\leavevmode\vadjust pre{\hypertarget{ref-carrizosa2021generating}{}}%
\CSLLeftMargin{{[}15{]} }%
\CSLRightInline{E. Carrizosa, J. Ramırez-Ayerbe, and D. Romero, {``Generating {Collective Counterfactual Explanations} in {Score-Based Classification} via {Mathematical Optimization},''} 2021.}

\leavevmode\vadjust pre{\hypertarget{ref-rabanser2019failing}{}}%
\CSLLeftMargin{{[}16{]} }%
\CSLRightInline{S. Rabanser, S. Günnemann, and Z. Lipton, {``Failing loudly: {An} empirical study of methods for detecting dataset shift,''} \emph{Advances in Neural Information Processing Systems}, vol. 32, 2019.}

\leavevmode\vadjust pre{\hypertarget{ref-chandola2009anomaly}{}}%
\CSLLeftMargin{{[}17{]} }%
\CSLRightInline{V. Chandola, A. Banerjee, and V. Kumar, {``Anomaly detection: {A} survey,''} \emph{ACM computing surveys (CSUR)}, vol. 41, no. 3, pp. 1--58, 2009.}

\leavevmode\vadjust pre{\hypertarget{ref-widmer1996learning}{}}%
\CSLLeftMargin{{[}18{]} }%
\CSLRightInline{G. Widmer and M. Kubat, {``Learning in the presence of concept drift and hidden contexts,''} \emph{Machine learning}, vol. 23, no. 1, pp. 69--101, 1996.}

\leavevmode\vadjust pre{\hypertarget{ref-gama2014survey}{}}%
\CSLLeftMargin{{[}19{]} }%
\CSLRightInline{J. Gama, I. Žliobaitė, A. Bifet, M. Pechenizkiy, and A. Bouchachia, {``A survey on concept drift adaptation,''} \emph{ACM computing surveys (CSUR)}, vol. 46, no. 4, pp. 1--37, 2014.}

\leavevmode\vadjust pre{\hypertarget{ref-nelson2015evaluating}{}}%
\CSLLeftMargin{{[}20{]} }%
\CSLRightInline{K. Nelson, G. Corbin, M. Anania, M. Kovacs, J. Tobias, and M. Blowers, {``Evaluating model drift in machine learning algorithms,''} in \emph{2015 {IEEE Symposium} on {Computational Intelligence} for {Security} and {Defense Applications} ({CISDA})}, 2015, pp. 1--8.}

\leavevmode\vadjust pre{\hypertarget{ref-ackerman2021machine}{}}%
\CSLLeftMargin{{[}21{]} }%
\CSLRightInline{S. Ackerman, P. Dube, E. Farchi, O. Raz, and M. Zalmanovici, {``Machine {Learning Model Drift Detection Via Weak Data Slices},''} in \emph{2021 {IEEE}/{ACM Third International Workshop} on {Deep Learning} for {Testing} and {Testing} for {Deep Learning} ({DeepTest})}, 2021, pp. 1--8.}

\leavevmode\vadjust pre{\hypertarget{ref-deoliveira2021framework}{}}%
\CSLLeftMargin{{[}22{]} }%
\CSLRightInline{R. M. B. de Oliveira and D. Martens, {``A framework and benchmarking study for counterfactual generating methods on tabular data,''} \emph{Applied Sciences}, vol. 11, no. 16, p. 7274, 2021.}

\leavevmode\vadjust pre{\hypertarget{ref-dombrowski2021diffeomorphic}{}}%
\CSLLeftMargin{{[}23{]} }%
\CSLRightInline{A.-K. Dombrowski, J. E. Gerken, and P. Kessel, {``Diffeomorphic explanations with normalizing flows,''} 2021.}

\leavevmode\vadjust pre{\hypertarget{ref-pindyck2014microeconomics}{}}%
\CSLLeftMargin{{[}24{]} }%
\CSLRightInline{R. S. Pindyck and D. L. Rubinfeld, \emph{Microeconomics}. {Pearson Education}, 2014.}

\leavevmode\vadjust pre{\hypertarget{ref-gretton2012kernel}{}}%
\CSLLeftMargin{{[}25{]} }%
\CSLRightInline{A. Gretton, K. M. Borgwardt, M. J. Rasch, B. Schölkopf, and A. Smola, {``A kernel two-sample test,''} \emph{The Journal of Machine Learning Research}, vol. 13, no. 1, pp. 723--773, 2012.}

\leavevmode\vadjust pre{\hypertarget{ref-berlinet2011reproducing}{}}%
\CSLLeftMargin{{[}26{]} }%
\CSLRightInline{A. Berlinet and C. Thomas-Agnan, \emph{Reproducing kernel {Hilbert} spaces in probability and statistics}. {Springer Science \& Business Media}, 2011.}

\leavevmode\vadjust pre{\hypertarget{ref-arcones1992bootstrap}{}}%
\CSLLeftMargin{{[}27{]} }%
\CSLRightInline{M. A. Arcones and E. Gine, {``On the bootstrap of {U} and {V} statistics,''} \emph{The Annals of Statistics}, pp. 655--674, 1992.}

\leavevmode\vadjust pre{\hypertarget{ref-hanneke2007bound}{}}%
\CSLLeftMargin{{[}28{]} }%
\CSLRightInline{S. Hanneke, {``A bound on the label complexity of agnostic active learning,''} in \emph{Proceedings of the 24th international conference on {Machine} learning}, 2007, pp. 353--360.}

\leavevmode\vadjust pre{\hypertarget{ref-innes2018fashionable}{}}%
\CSLLeftMargin{{[}29{]} }%
\CSLRightInline{M. Innes \emph{et al.}, {``Fashionable modelling with flux,''} 2018. Available: \url{https://arxiv.org/abs/1811.01457}}

\leavevmode\vadjust pre{\hypertarget{ref-lakshminarayanan2016simple}{}}%
\CSLLeftMargin{{[}30{]} }%
\CSLRightInline{B. Lakshminarayanan, A. Pritzel, and C. Blundell, {``Simple and scalable predictive uncertainty estimation using deep ensembles,''} 2016. Available: \url{https://arxiv.org/abs/1612.01474}}

\leavevmode\vadjust pre{\hypertarget{ref-kaggle2011give}{}}%
\CSLLeftMargin{{[}31{]} }%
\CSLRightInline{Kaggle, {``Give me some credit, {Improve} on the state of the art in credit scoring by predicting the probability that somebody will experience financial distress in the next two years.''} {Kaggle}, 2011. Available: \url{https://www.kaggle.com/c/GiveMeSomeCredit}}

\leavevmode\vadjust pre{\hypertarget{ref-yeh2009comparisons}{}}%
\CSLLeftMargin{{[}32{]} }%
\CSLRightInline{I.-C. Yeh and C. Lien, {``The comparisons of data mining techniques for the predictive accuracy of probability of default of credit card clients,''} \emph{Expert systems with applications}, vol. 36, no. 2, pp. 2473--2480, 2009.}

\leavevmode\vadjust pre{\hypertarget{ref-pedregosa2011scikitlearn}{}}%
\CSLLeftMargin{{[}33{]} }%
\CSLRightInline{F. Pedregosa \emph{et al.}, {``Scikit-learn: {Machine} learning in {Python},''} \emph{the Journal of machine Learning research}, vol. 12, pp. 2825--2830, 2011.}

\leavevmode\vadjust pre{\hypertarget{ref-pace1997sparse}{}}%
\CSLLeftMargin{{[}34{]} }%
\CSLRightInline{R. K. Pace and R. Barry, {``Sparse spatial autoregressions,''} \emph{Statistics \& Probability Letters}, vol. 33, no. 3, pp. 291--297, 1997.}

\leavevmode\vadjust pre{\hypertarget{ref-carlisle2019racist}{}}%
\CSLLeftMargin{{[}35{]} }%
\CSLRightInline{M. Carlisle, {``Racist data destruction? - a {Boston} housing dataset controversy,''} 2019, Available: \url{https://medium.com/@docintangible/racist-data-destruction-113e3eff54a8}}

\leavevmode\vadjust pre{\hypertarget{ref-miller2020strategic}{}}%
\CSLLeftMargin{{[}36{]} }%
\CSLRightInline{J. Miller, S. Milli, and M. Hardt, {``Strategic {Classification} is {Causal Modeling} in {Disguise},''} in \emph{Proceedings of the 37th {International Conference} on {Machine Learning}}, Nov. 2020, pp. 6917--6926. Accessed: Nov. 03, 2022. {[}Online{]}. Available: \url{https://proceedings.mlr.press/v119/miller20b.html}}

\leavevmode\vadjust pre{\hypertarget{ref-barocas2022fairness}{}}%
\CSLLeftMargin{{[}37{]} }%
\CSLRightInline{S. Barocas, M. Hardt, and A. Narayanan, {``Fairness and machine learning,''} Dec. 23, 2022. \url{https://fairmlbook.org/index.html} (accessed Nov. 08, 2022).}

\leavevmode\vadjust pre{\hypertarget{ref-borisov2021deep}{}}%
\CSLLeftMargin{{[}38{]} }%
\CSLRightInline{V. Borisov, T. Leemann, K. Seßler, J. Haug, M. Pawelczyk, and G. Kasneci, {``Deep neural networks and tabular data: {A} survey,''} 2021. Available: \url{https://arxiv.org/abs/2110.01889}}

\leavevmode\vadjust pre{\hypertarget{ref-grinsztajn2022why}{}}%
\CSLLeftMargin{{[}39{]} }%
\CSLRightInline{L. Grinsztajn, E. Oyallon, and G. Varoquaux, {``Why do tree-based models still outperform deep learning on tabular data?''} 2022. Available: \url{https://arxiv.org/abs/2207.08815}}

\end{CSLReferences}

\newpage

\hypertarget{appendix}{%
\section*{Appendix}\label{appendix}}
\addcontentsline{toc}{section}{Appendix}

Granular results for all of our experiments can be found in this online companion: \url{https://www.paltmeyer.com/endogenous-macrodynamics-in-algorithmic-recourse/}. The Github repository containing all the code used to produce the results in this paper can be found here: \url{https://github.com/pat-alt/endogenous-macrodynamics-in-algorithmic-recourse}.

\end{document}